\documentclass[10pt,twocolumn,letterpaper]{article}

\usepackage{cvpr}              

\usepackage{graphicx}
\usepackage{amsmath}
\usepackage{amssymb}
\usepackage{booktabs}
\usepackage{multirow}
\usepackage{algorithm}
\usepackage{algpseudocode}

\usepackage[pagebackref,breaklinks,colorlinks]{hyperref}
\usepackage{array}

\usepackage[capitalize]{cleveref}
\crefname{section}{Sec.}{Secs.}
\Crefname{section}{Section}{Sections}
\Crefname{table}{Table}{Tables}
\crefname{table}{Tab.}{Tabs.}

\newcommand{\myparagraph}[1]{\smallskip\noindent\textbf{#1}}

\newcommand{\tabincell}[2]{\begin{tabular}{@{}#1@{}}#2\end{tabular}}
\def\cvprPaperID{6862} 
\def\confName{CVPR}
\def\confYear{2023}

\begin{document}

\title{Backdoor Cleansing with Unlabeled Data}

\author{Lu Pang, Tao Sun, Haibin Ling, Chao Chen\\
Stony Brook University\\
{\tt\small \{luppang,tao,hling\}@cs.stonybrook.edu, 
chao.chen.1@stonybrook.edu}}
\maketitle

\begin{abstract}
Due to the increasing computational demand of Deep Neural Networks (DNNs), companies and organizations have begun to outsource the training process. However, the externally trained DNNs can potentially be backdoor attacked. 
It is crucial to defend against such attacks, \ie, to postprocess a suspicious model so that its backdoor behavior is mitigated while its normal prediction power on clean inputs remain uncompromised.
To remove the abnormal backdoor behavior, existing methods mostly rely on additional labeled clean samples. However, such requirement may be unrealistic as the training data are often unavailable to end users. 
In this paper, we investigate the possibility of circumventing such barrier. We propose a novel defense method that does not require training labels.
Through a carefully designed layer-wise weight re-initialization and knowledge distillation, our method can effectively cleanse backdoor behaviors of a suspicious network {with negligible compromise in} its normal behavior. 
In experiments, we show that our method, trained without labels, is on-par with state-of-the-art defense methods trained using labels. We also observe promising defense results even on out-of-distribution data. This makes our method very practical. Code is available at: \url{https://github.com/luluppang/BCU}.


\end{abstract}

\section{Introduction}
\label{sec:intro}
Deep Neural Networks (DNNs) have achieved impressive performance in many tasks, \textit{e.g.}, image classification~\cite{deng2009imagenet}, 3D point cloud generation~\cite{luo2021diffusion} and object tracking~\cite{zheng2021improving}. However, the success usually relies on abundant training data and computational resources. Companies and organizations thus often outsource the training process to cloud computing or utilize pretrained models from third-party platforms. Unfortunately, the untrustworthy providers may potentially introduce backdoor attacks to the externally trained DNNs~\cite{gu2019badnets, liutrojaning}. During the training stage of a backdoor attack, an adversary stealthily injects a small portion of poisoned training data to associate a particular trigger with target class labels. During the inference stage, backdoor models predict accurately on clean samples but misclassify samples with triggers to the target class. Common triggers include  black-white checkerboard~\cite{gu2019badnets}, random noise pattern~\cite{chen2017blended}, physical object~\cite{wenger2021backdoor}, \etc. 

\begin{figure}
    \centering
    \includegraphics[width=0.45\textwidth]{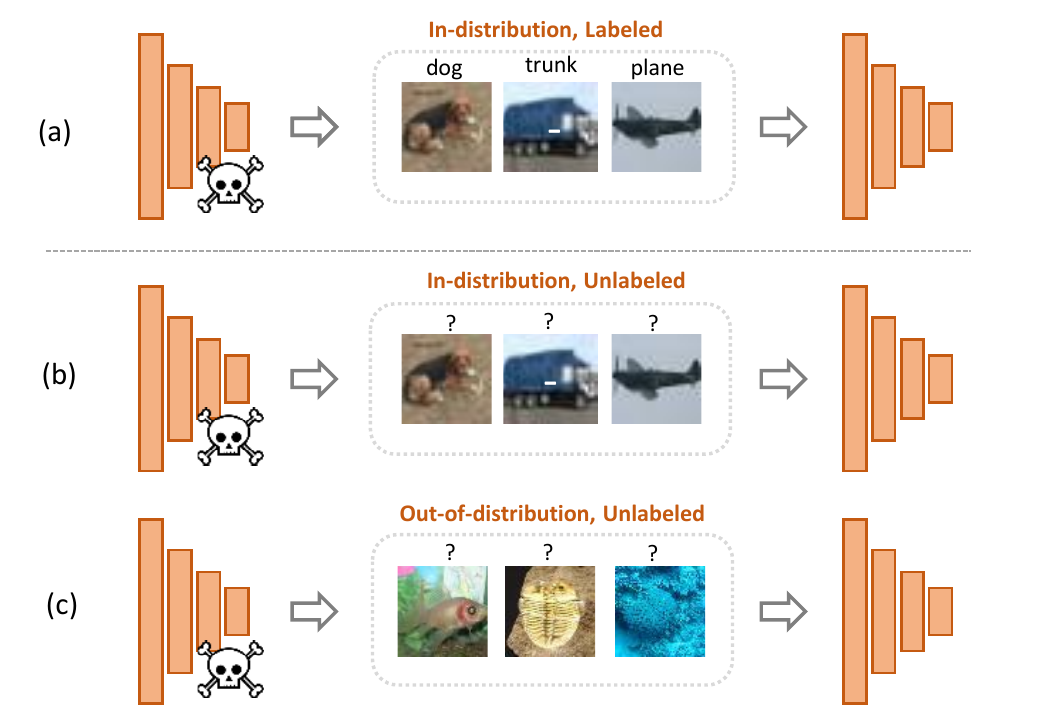}
    \caption{(a) Previous works use labeled in-distribution data to cleanse backdoor. Our work uses unlabeled in-distribution  (b) or out-of-distribution data (c).}
    \vspace{-3.5mm}
    \label{fig:unlabeled}
\end{figure}

To defend against backdoor attacks, one needs to post-process a suspicious model so that its backdoor behavior is mitigated, and meanwhile, its normal prediction power on clean inputs remains uncompromised. 
To remove the abnormal backdoor behavior, existing methods mostly rely on additional labeled in-distribution clean samples ~\cite{li2021NAD, liu2018fine-pruning, wu2021ANP, xia2022NA-RGD, zeng2021i-bau, zhao2020MCR}. 
For example, Fine-Pruning~\cite{liu2018fine-pruning} first prunes the dormant neurons for clean samples and then finetunes the model using ground-truth labels. Neural Attention Distillation (NAD)~\cite{li2021NAD}, a knowledge distillation-based method, uses labeled clean data to supervise the learning of a student model. Adversarial Neuron Pruning (ANP)~\cite{wu2021ANP} learns a mask to prune sensitive neurons with labeled clean data. 
These methods require $1\%-5\%$ labeled clean training samples to effectively remove backdoor. Such requirement, however, is unrealistic in practice as the training data are often unavailable to end-users. 


In this paper, we explore the possibility of circumventing such barrier with unlabeled data. As shown in Figure~\ref{fig:unlabeled}, we propose a novel defense method that does not require training labels. Meanwhile, we explore the ambitious goal of using only out-of-distribution data. These goals 
make the proposed defense method much more practical. End-users can be completely agnostic of the training set. To run the defense algorithm, they only need to collect some unlabeled data that do not have to resemble the training samples.

Inspired by knowledge distillation~\cite{gou2021knowledge}, we use a student model to acquire benign knowledge from a suspicious teacher model through their predictions on the readily available unlabeled data. Since the unlabeled data are usually clean images or images with slightly random noise, they are distinct from poisoned images with triggers. Therefore, trigger-related behaviors will not be evoked during the distillation. This effectively cleanses backdoor behaviors without significantly compromising the model's normal behavior. To ensure the student model focuses on the benign knowledge, which can be layer dependent, we propose an adaptive layer-wise weight re-initialization for the student model. Empirically, we demonstrate that even without labels, the proposed method can still successfully defend against the backdoor attacks. We also observe very promising defense results even with out-of-distribution unlabeled data that do not belong to the original training classes. 

Our contributions are summarized as follows: 
\begin{enumerate}
\vspace{-1.9mm}\item For the first time, we propose to defend against backdoor attacks using unlabeled data. This provides a practical solution to end-users under threat.
\vspace{-1.9mm}\item We devise a framework with knowledge distillation to transfer normal behavior of a suspicious teacher model to a student model while cleansing backdoor behaviors. Since the normal/backdoor knowledge can be layer-dependent, we design an adaptive layer-wise initialization strategy for the student model. 
\vspace{-1.9mm}\item Extensive experiments are conducted on two benchmark datasets, CIFAR10~\cite{krizhevsky2009cifar10} and GTSRB~\cite{stallkamp2012gtsrb}. Our method, trained without labels, is on-par with state-of-the-art defense methods trained with labels.
\vspace{-1.9mm}\item Meanwhile, we carry out an empirical study with out-of-distribution data. Our method achieves satisfactory defense performance against a majority of attacks. This sheds lights on a promising practical solution for end-users: they can use any collected images to cleanse a suspicious model. 
\end{enumerate}

\section{Related Work}
\label{sec:related_work}
\subsection{Backdoor Attack}
During a backdoor attack, the adversary embeds a trigger into a DNN model by poisoning a portion of the training dataset at the training stage. At the inference stage, the backdoor model classifies clean samples accurately while predicts backdoor samples as the target label. The poisoned training samples are attached with a specific trigger and relabeled as the target label. A simple trigger can be a black-white checkerboard~\cite{gu2019badnets} or a single pixel~\cite{tran2018spectral}. These triggers are not stealthy since they can be perceived by human eyes. More complex triggers are developed such as a sinusoidal strip~\cite{barni2019sig}, an input-aware dynamic pattern~\cite{nguyen2020IAB, salem2022dynamic}, \etc. Recent works~\cite{wang2022bppattack, doan2021lira, nguyen2021wanet, Liu2020Refool} design more imperceptible triggers. Refool~\cite{Liu2020Refool} utilizes a natural reflection phenomenon to design triggers. WaNet~\cite{nguyen2021wanet} uses elastic image warping technique to generate triggers. Lira~\cite{doan2021lira} jointly optimizes trigger injection function and classification loss function to get stealthy triggers. BppAttack~\cite{wang2022bppattack} improves the quality of triggers by using image quantization and injects triggers effectively with contrastive adversarial learning. 
Besides, some methods~\cite{barni2019sig, turner2019lc, shafahi2018poisonfrog, saha2020hidden} keep the original label of poisoned samples same as the target label. Such clean-label setting is more imperceptive for human inspectors. The key of these methods is to make models misclassify the clean target-label samples during the training process. Also, recent works show that backdoor attacks can be applied to federated learning~\cite{xie2021crfl}, transfer learning~\cite{saha2020hidden}, self-supervised learning~\cite{saha2022backdoor}, 3D point cloud classification~\cite{xiang2021backdoor}, visual object tracking~\cite{li2022few} and crowd counting~\cite{sun2022backdoor}.

\subsection{Backdoor Defense}

\begin{figure*}
    \centering
    \includegraphics[width=0.9\textwidth]{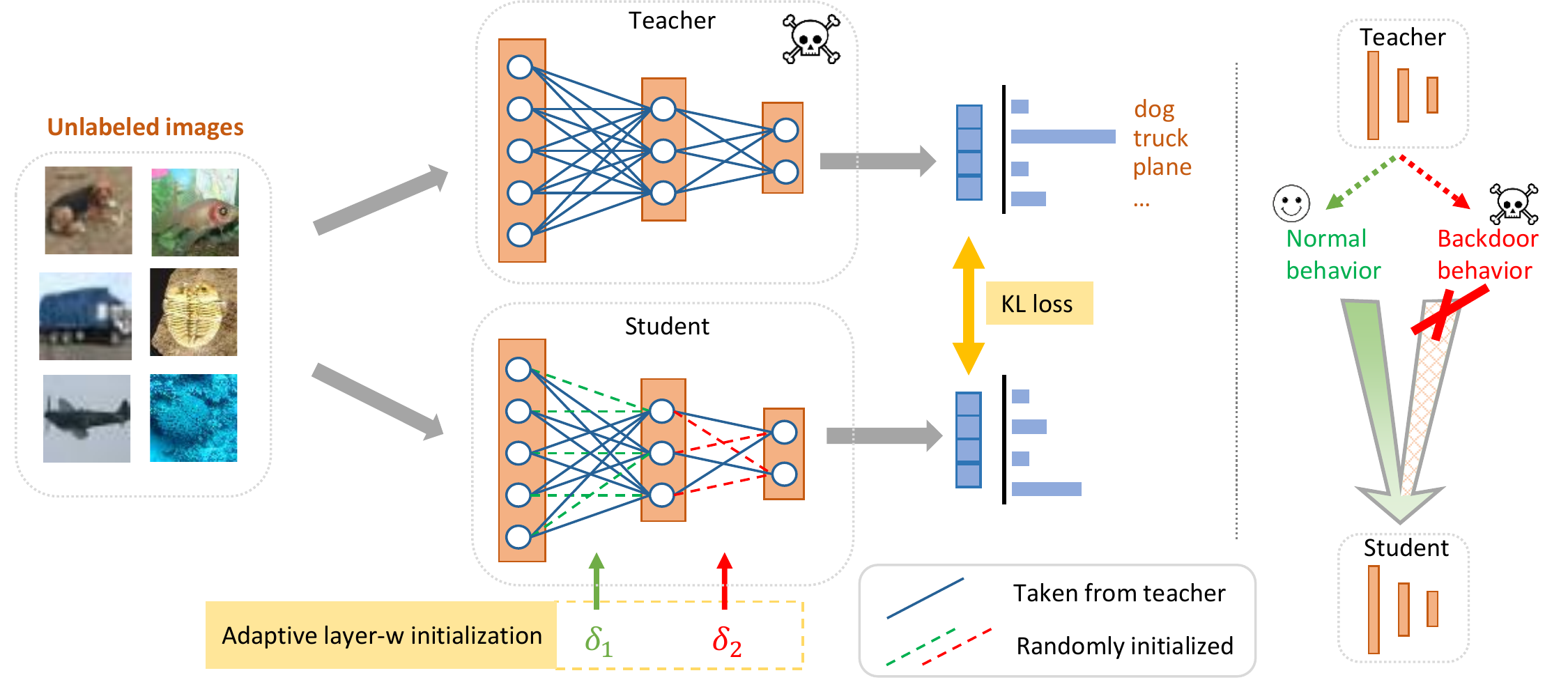}
    \caption{Proposed backdoor cleansing framework. The student model learns normal behavior from the teacher model through knowledge distillation on unlabeled images. Backdoor behavior of the teacher model is neglected.}
    \vspace{-2mm}
    \label{fig:framework}
\end{figure*}

In model reconstruction-based defense, given a trained suspicious model, defenders modify the model directly to eliminate backdoor effects. Most methods in this category first synthesize possible triggers and then utilize synthesized triggers to mitigate backdoor effects~\cite{wang2019NC, zhu2020gangsweep, guan2022shapley, chen2022quarantine, chen2019deepinspect}. With some clean samples, Neural Cleanse (NC)~\cite{wang2019NC} synthesizes a trigger for each class and uses a Median Absolute Deviation (MAD) outlier detection algorithm to detect the final trigger. Then, an unlearning strategy is designed to unlearn the backdoor effects. Following NC~\cite{wang2019NC}, other methods~\cite{chen2019deepinspect, zhu2020gangsweep, chen2022quarantine, guan2022shapley} are proposed to improve the quality of synthesized triggers. For example, ShapPruning~\cite{guan2022shapley} employs Shapley estimation to synthesize triggers and then detect sensitive neurons to synthesized triggers. Chen et al.~\cite{chen2022quarantine} locates a ``wining backdoor lottery ticket'' to preserve trigger-related information. These methods heavily depend on the quality of the synthesized triggers, and thus can be unsatisfactory when facing more advanced triggers~\cite{salem2022dynamic,doan2021lira}.

Other works explore pruning-based defense methods~\cite{liu2018fine-pruning, wu2021ANP}. The core idea is to detect and prune bad neurons. For example, Fine-Pruning~\cite{liu2018fine-pruning} prunes bad neurons of the last convolution layer, and then uses clean samples to finetune the pruned model. Adversarial Neuron Pruning (ANP)~\cite{wu2021ANP} treats pruning sensitive neurons as a minimax problem under adversarial neuron perturbations.  The Implicit Backdoor Adversarial Unlearning (I-BAU) algorithm~\cite{zeng2021i-bau} solves the minimax optimization by utilizing the implicit hyper-gradient. Besides, Mode Connectivity Repair (MCR)~\cite{zhao2020MCR} is explored to remove backdoor effects. 
%
Although effective, these methods require labeled clean samples, which in practice may not be available. By contrast, our solution, also in the model-reconstructing category, does not need labeled clean samples.

Knowledge distillation has been used in backdoor mitigation~\cite{li2021NAD, xia2022NA-RGD}. Both Neural Attention Distillation (NAD)~\cite{li2021NAD} and Attention Relation Graph Distillation (ARGD)~\cite{xia2022NA-RGD} transfer feature attention knowledge of a finetuned backdoor model into the original backdoor model. These methods crucially rely on the finetuning stage, and thus depend on labeled clean samples.
The key insight of our method is that model prediction on data automatically carries rich and benign knowledge of the original model. Through a layer-adaptive weight initialization strategy, \emph{our method can directly cleanse backdoors without any label}.

While most existing works assume the defense as a postprocessing step, we also note some recent methods focusing on designing backdoor-resilient training strategy. Since a defender can access the training process, some works modify the training strategy to train a robust model~\cite{huang2021DBD, li2021anti}. Huang et al.~\cite{huang2021DBD} decompose end-to-end training process into three stages including self-supervised feature learning, classifier learning and finetuning whole classifier with filtered samples. Based on the characteristics of backdoor model training, Li et al.~\cite{li2021anti} proposes a two-stage gradient ascent strategy instead of standard training. Other studies mitigate backdoor effects via randomized smoothing~\cite{muravev2021certified}, noise injection~\cite{ma2019data} and strong data augmentation~\cite{borgnia2021strong}, \etc. 

\begin{figure*}[!t]
	\centering	
	\begin{subfigure}[]{0.195\textwidth}
		\includegraphics[width=\textwidth]{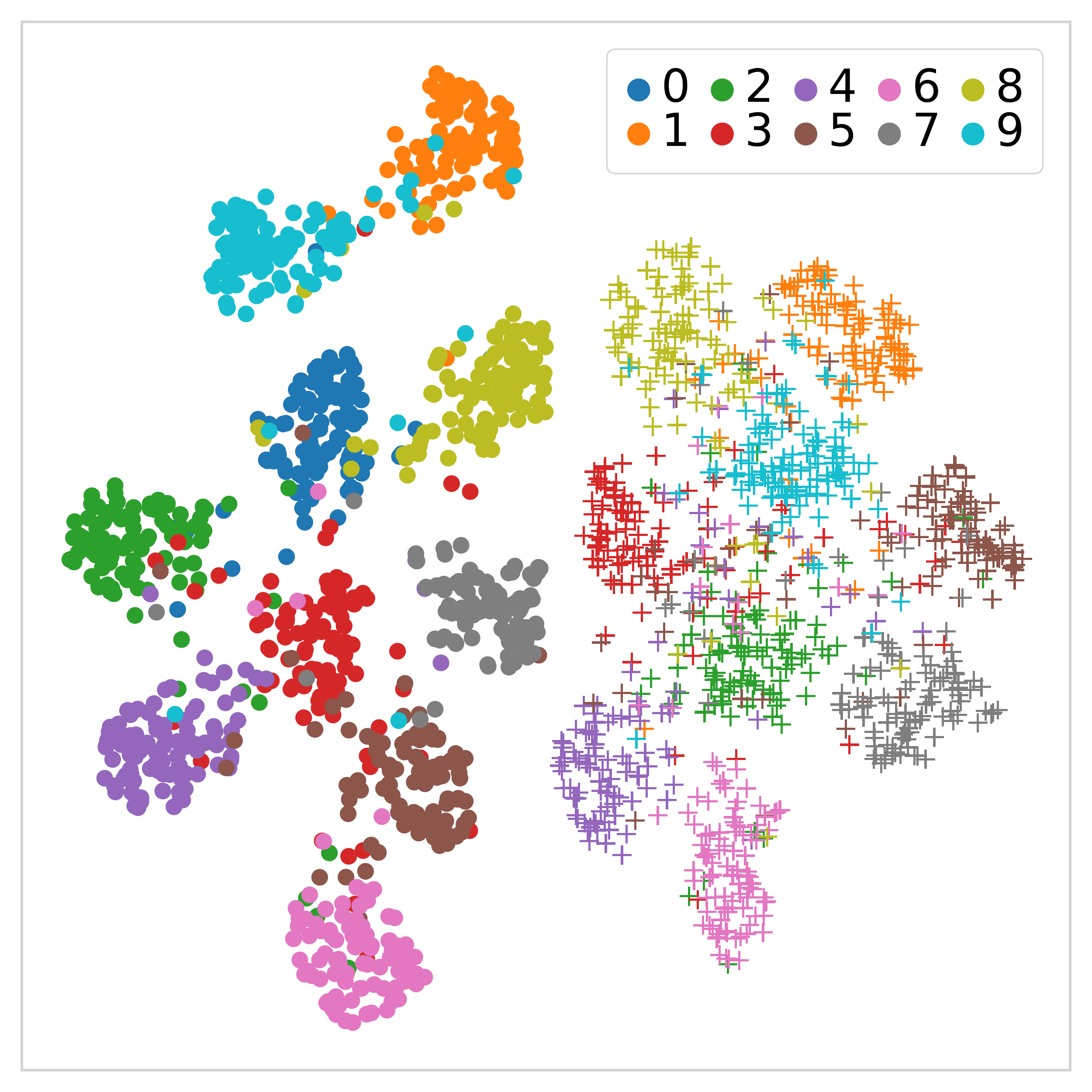}
		\caption{Teacher}	
		\label{fig:cifar_tsne:a}
	\end{subfigure}
	\begin{subfigure}[]{0.195\textwidth}
		\includegraphics[width=\textwidth]{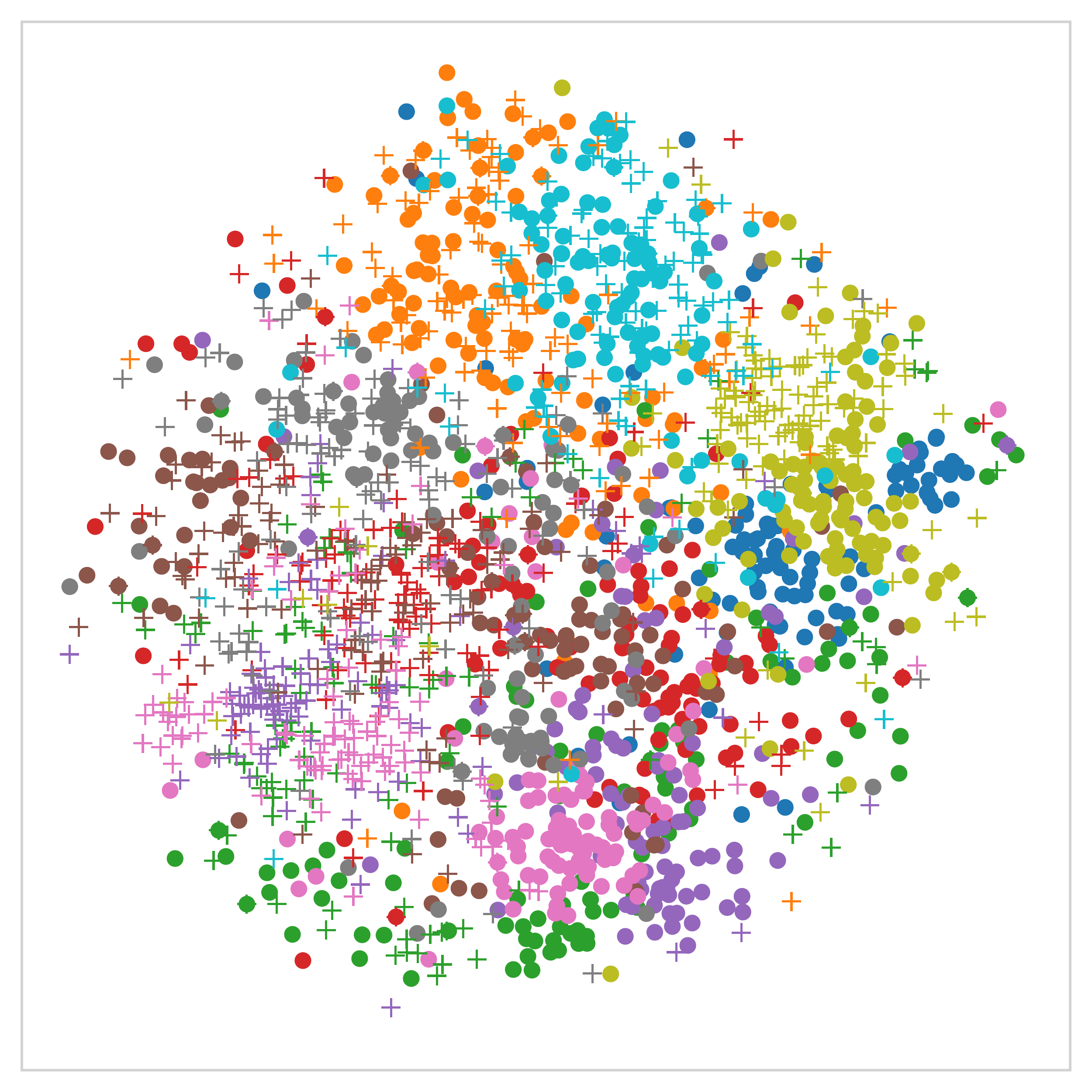}
		\caption{Stu (adaptive init, ep 0)}	
		\label{fig:cifar_tsne:b}
	\end{subfigure}
	\begin{subfigure}[]{0.195\textwidth}
		\includegraphics[width=\textwidth]{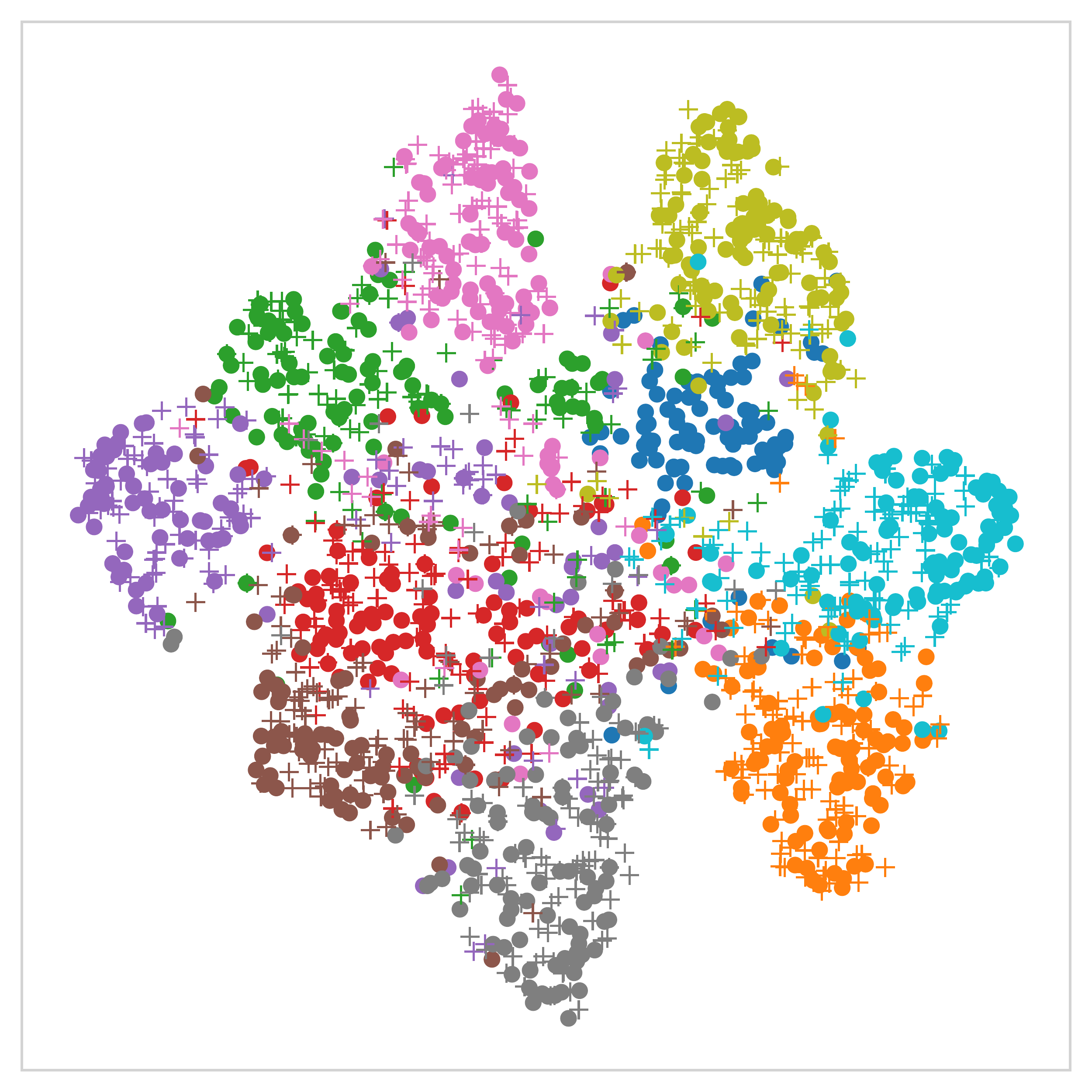}
		\caption{Stu (adaptive init, ep 1)}	
		\label{fig:cifar_tsne:c}
	\end{subfigure}
	\begin{subfigure}[]{0.195\textwidth}
		\includegraphics[width=\textwidth]{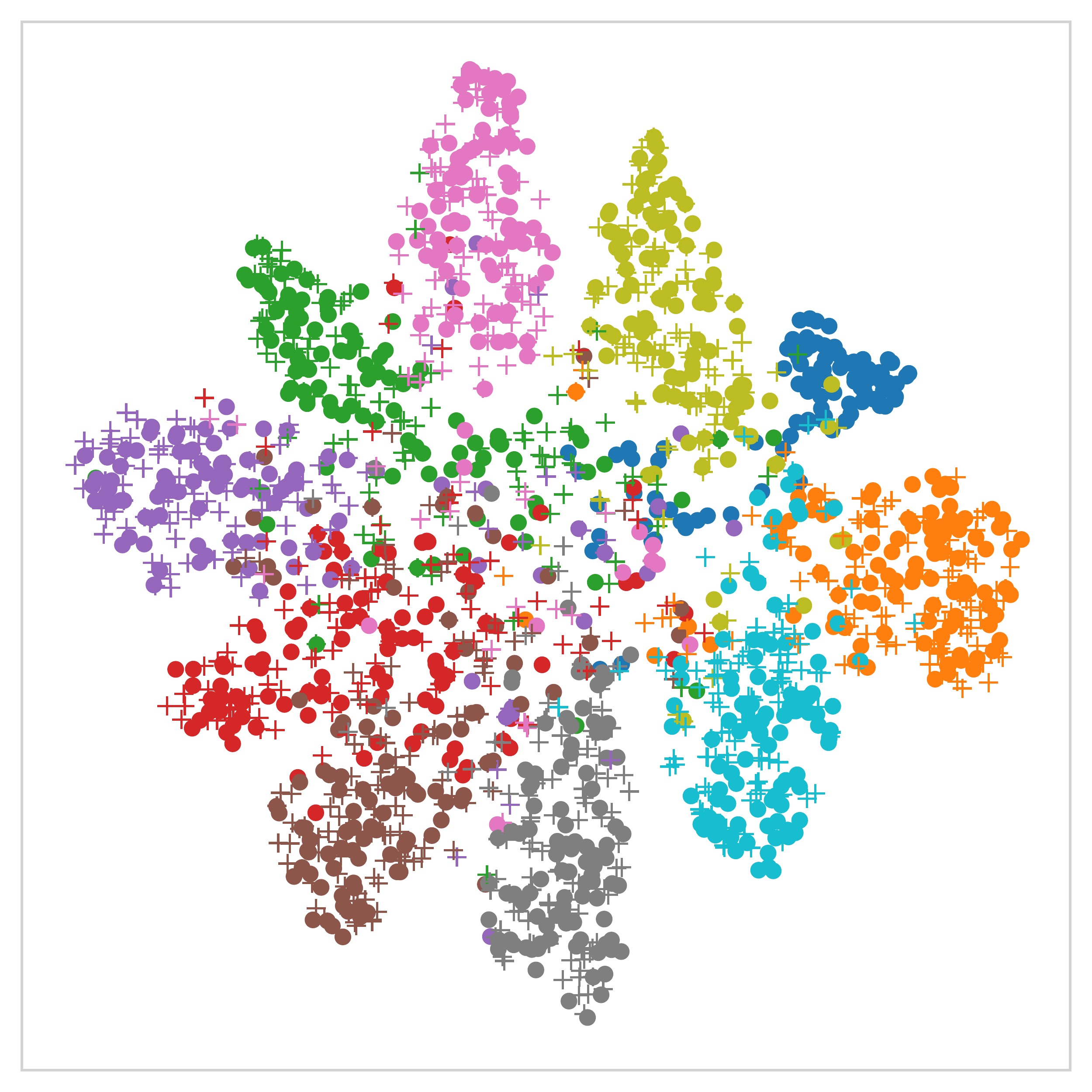}
		\caption{Stu (adaptive init, ep 2)}	
		\label{fig:cifar_tsne:d}
	\end{subfigure}
		\begin{subfigure}[]{0.195\textwidth}
		\includegraphics[width=\textwidth]{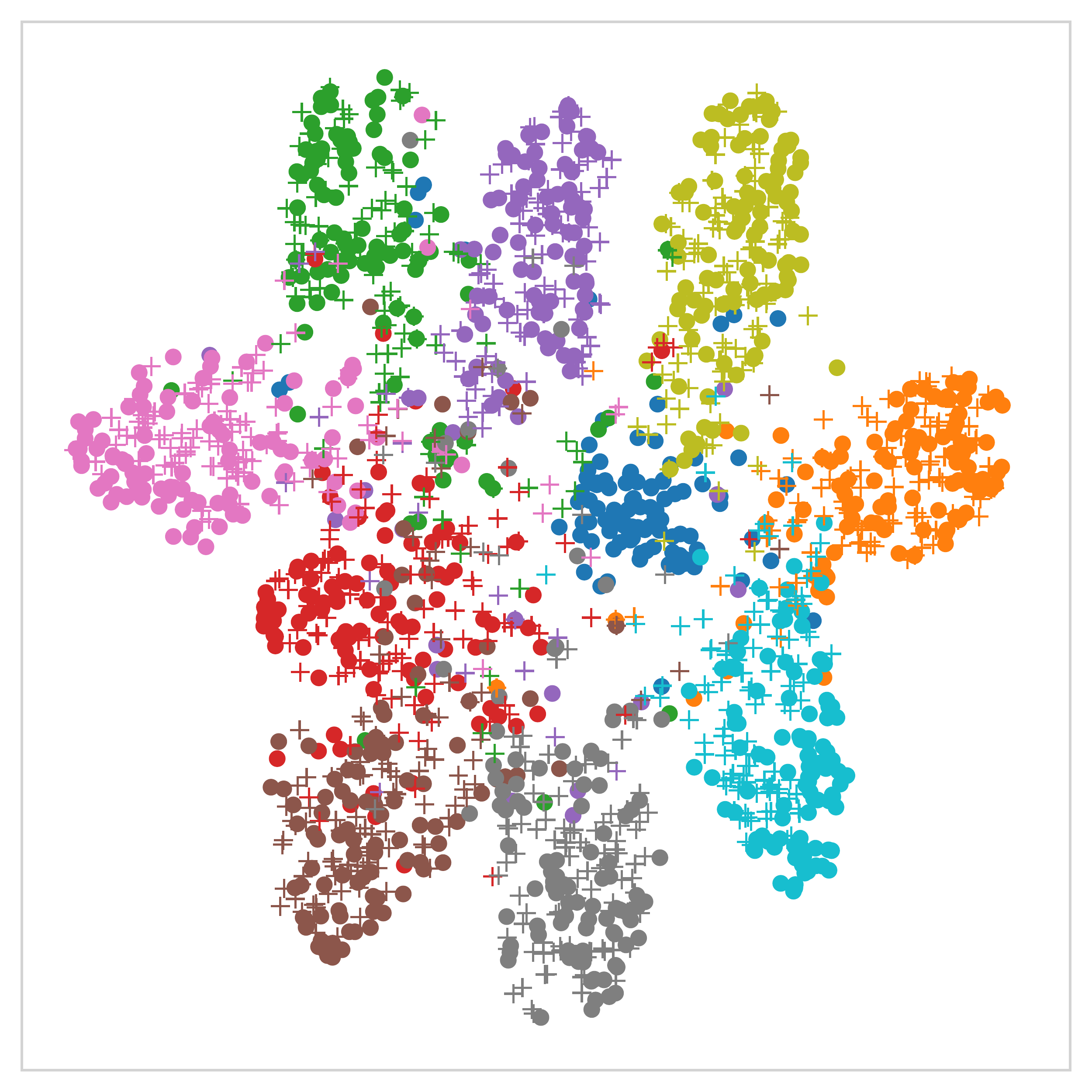}
		\caption{Stu (adaptive init, ep 11)}	
		\label{fig:cifar_tsne:e}
	\end{subfigure}
	
	\begin{subfigure}[]{0.195\textwidth}
	\includegraphics[width=\textwidth]{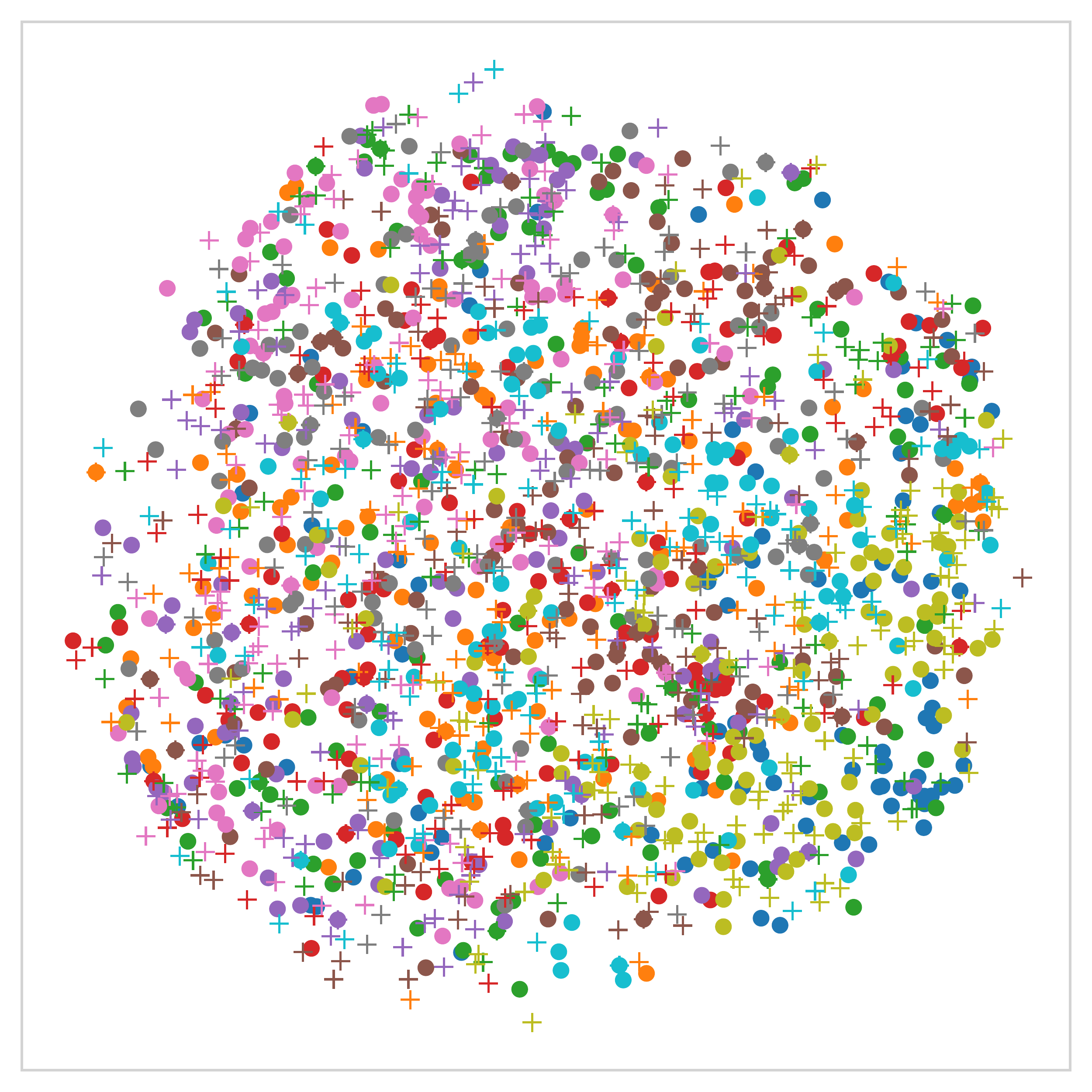}
	\caption{Stu (uniform init, ep 0)}
	\label{fig:cifar_tsne:f}
 	\end{subfigure}
	\begin{subfigure}[]{0.195\textwidth}
	\includegraphics[width=\textwidth]{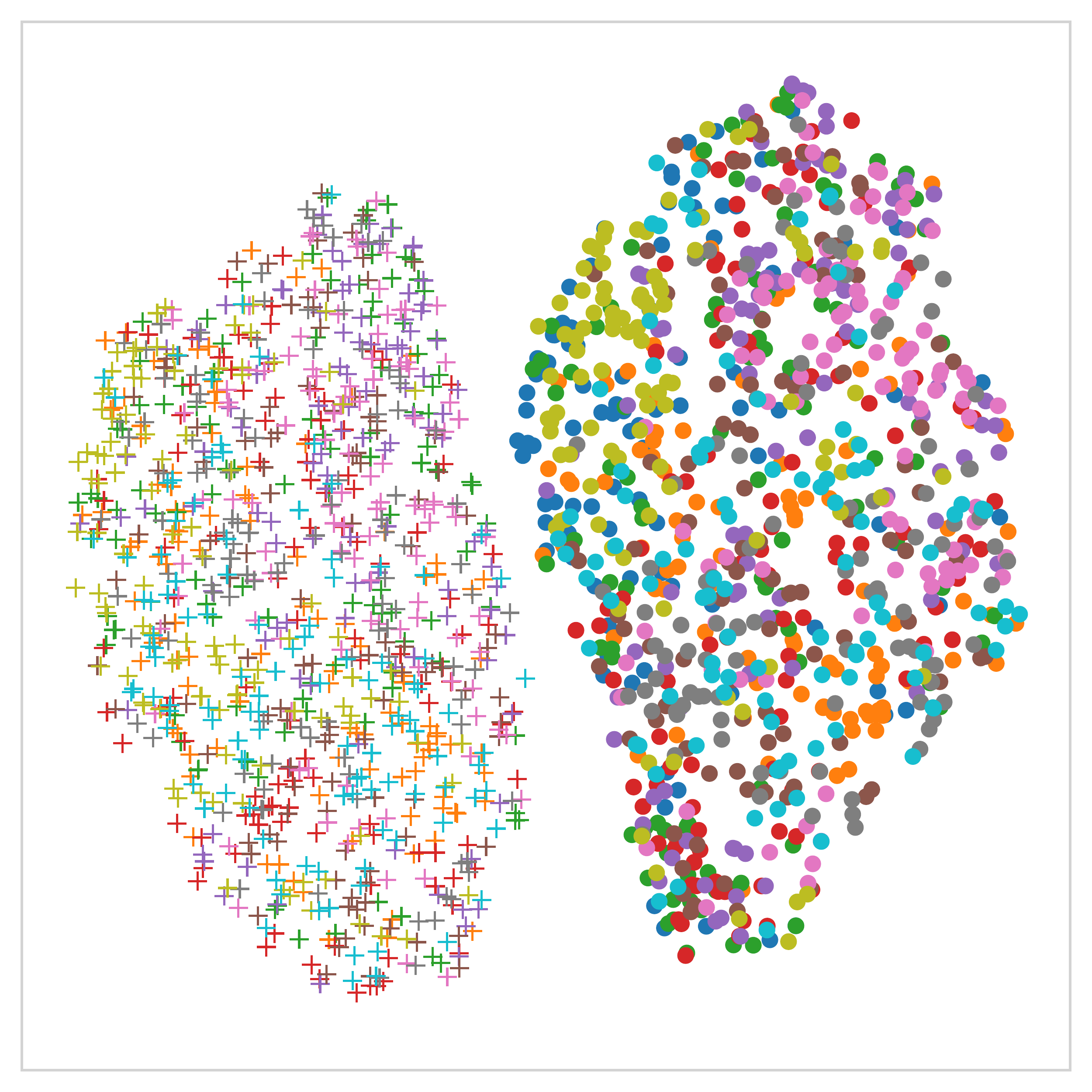}
	\caption{Stu (1st layer init, ep 0)}	
	\label{fig:cifar_tsne:g}
 	\end{subfigure}
 	\begin{subfigure}[]{0.195\textwidth}
	\includegraphics[width=\textwidth]{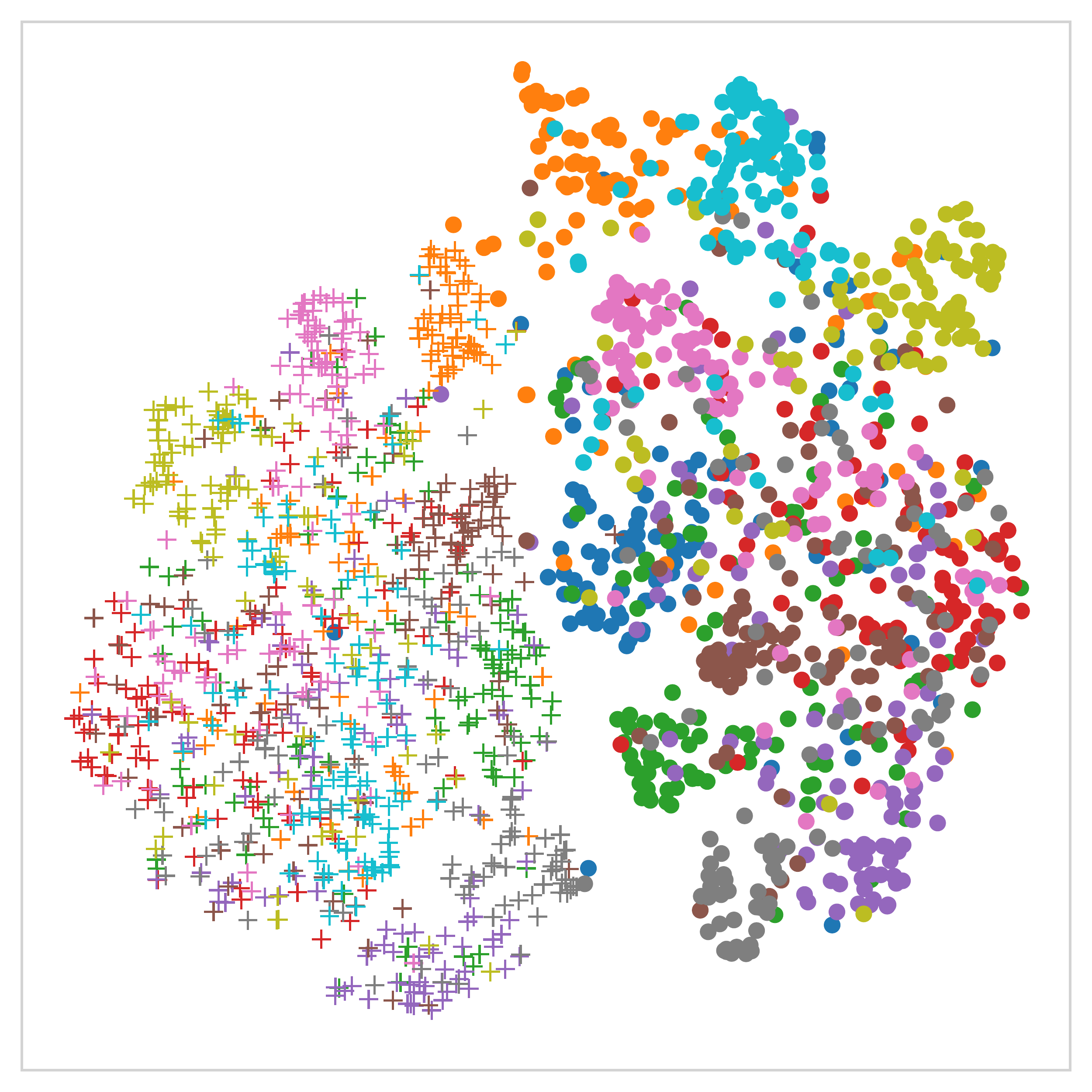}
	\caption{Stu (2nd layer init, ep 0)}	
	\label{fig:cifar_tsne:h}
 	\end{subfigure}
 	\begin{subfigure}[]{0.195\textwidth}
	\includegraphics[width=\textwidth]{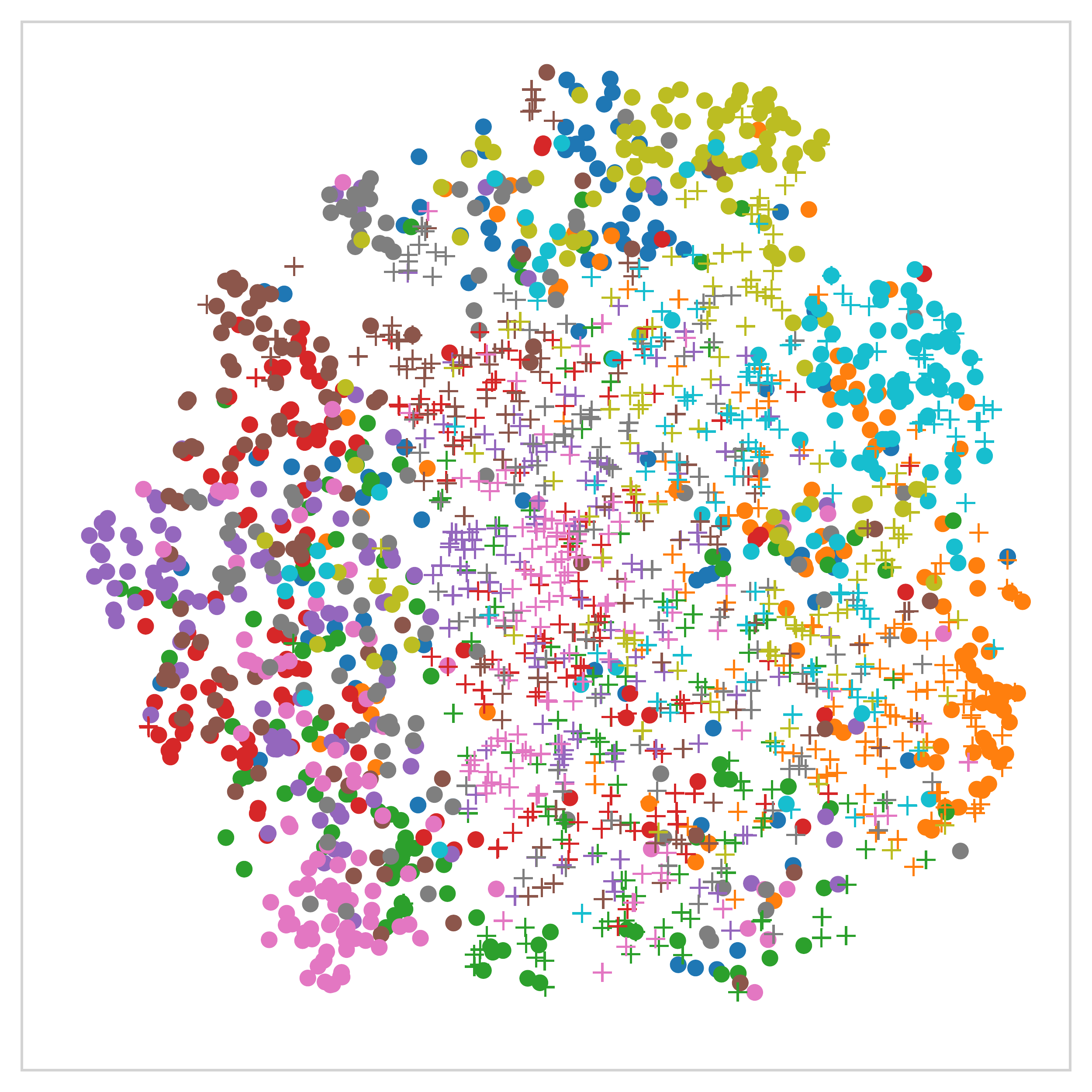}
    \caption{Stu (3rd layer init, ep 0)}	
	\label{fig:cifar_tsne:i}
 	\end{subfigure}
 	\begin{subfigure}[]{0.195\textwidth}
	\includegraphics[width=\textwidth]{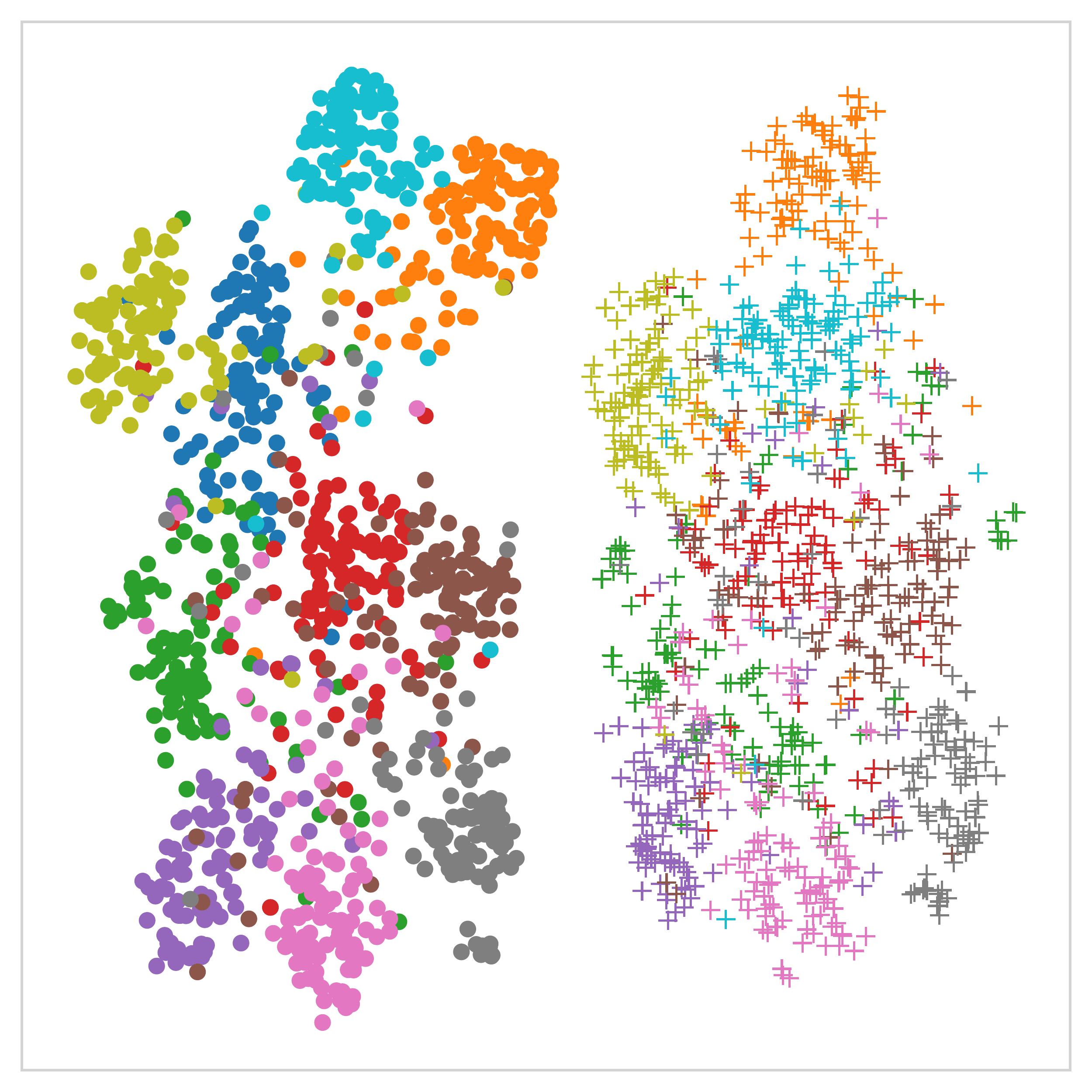}
    \caption{Stu (4th layer init, ep 0)}
	\label{fig:cifar_tsne:j}
 	\end{subfigure}
 	
    \caption{$t$-SNE visualization of penultimate features on CIFAR10 from \textit{Badnets} attack. \textbf{Top}: the teacher model and student models at different training epochs  with adaptive layer-wise initialization. \textbf{Bottom}: student models at epoch 0 with different initialization strategies. Each color denotes a class. `$\circ$' are clean images and `$+$' the corresponding backdoor ones. More discussions can be found in Section \ref{sec:method:initialization}.}
    \label{fig:cifar_tsne}
\end{figure*}

\section{Method}
\label{sec:method}

Our main idea is to directly use knowledge distillation to cleanse backdoor behaviors. The rationale is three-folds. First, knowledge distillation directly transfers knowledge through the logits output, which carries the rich posterior probability distribution information of a model. By approximating the logits output on samples, the student model can naturally mimic the normal behavior of the teacher model. Second, we argue that the backdoor behavior is an abnormal phenomenon forced into the teacher model. Knowledge distillation through clean samples will implicitly regularize the transferred knowledge, and ``smooth'' out the abnormal behavior. Finally, prior study has observed that backdoor behavior is embodied in certain neurons whose distribution is layer dependent~\cite{lyu2022study}. By designing an adaptive weight initialization, we can more effectively transfer normal knowledge of the teacher model and filter out backdoor behavior. The framework of our method is illustrated in Figure \ref{fig:framework}.



\subsection{Preliminary}
\vspace{-2mm}
\myparagraph{Attack Setting.}
In backdoor attack for classification task, a DNN model $f_{\theta}: \mathcal{X}\longrightarrow \mathcal{Y}$ is trained, where $\mathcal{X} \subset \mathbb{R}^d$ is the input space and $\mathcal{Y} = \{1, 2, ..., K\}$ is the label space. An image dataset $D_{\rm attack} =\{(x_i, y_i) \in \mathcal{X}\times\mathcal{Y}\}_{i=1}^n$ is split by $D_{\rm attack}=D_{\rm clean} \bigcup D_{\rm backdoor}$, where $D_{\rm backdoor}$ is used to create backdoor images. The backdoor injection rate is defined as $\gamma = \frac{|D_{\rm backdoor}|}{|D_{\rm attack}|}$.  An image transformation function $\Phi(\cdot)$ transforms a clean image into a backdoor image, \textit{e.g.}, through stacking a checkerboard pattern to the original image. $\eta(\cdot,\cdot)$ transforms its ground truth label into a target label. The objective function for backdoor attack is 
\vspace{-1mm}
\begin{equation}
    \begin{aligned}
    \mathcal{L}_{\rm attack}  = & \mathbb{E}_{(x,y)\sim D_{\rm clean}} [\ell_{\rm ce}(f_\theta(x), y)] \\
    & + \mathbb{E}_{(x,y)\sim D_{\rm backdoor}} [\ell_{\rm ce}(f_\theta(\Phi(x)), \eta(x,y))] 
   \end{aligned}
\end{equation}
where $\ell_{\rm ce}$ is the cross entropy loss function. With this loss function, the obtained backdoor model is expected to behave normally on clean test images, while misclasify backdoor images to the target class label.

\myparagraph{Defense Setting.} We assume that defenders download a backdoored model from an untrustworthy platform and can not access the training process. Some clean images $D_{\rm defense}$ are given for backdoor defense. The goal of defense is to preserve the classification accuracy (ACC) on clean data and decrease the classification accuracy on backdoor images \ie attack success rate (ASR). 


\subsection{Backdoor Cleansing via Knowledge Distillation}
\label{sec:method:KD}
Our motivation is to directly extract clean information (or knowledge) from a suspicious model. Since a backdoor model usually behaves differently for clean and backdoor images, the trigger-related behaviors will not be evoked when the model is fed with clean images. Inspired by response-based knowledge distillation~\cite{hinton2015distilling}, we adopt the teacher-student framework to distillate benign knowledge from a suspicious teacher model through its predictions on clean images. As illustrated in Figure \ref{fig:framework}, the normal behaviors of the teacher model are transferred to the student model, while the backdoor behaviors are neglected. This effectively cleanses backdoor behaviors without significantly compromising the model’s performances on clean images. 

Since we use the the logits output of the teacher model as the supervision, our proposed framework does not need ground-truth labels. In fact, even when the input images are out-of-distribution data that do not belong to the training classes, the student model can acquire useful knowledge from the teacher model's predicted probabilities. 

Let $z^t$ and $z^s$ be the output logits of the teacher model and student model, respectively. Their temperature scaled probability vectors can be obtained as $p^t_{T}[k] = \frac{\exp(z_k^t/T)}{\sum_{j} \exp(z_j^t/T)}$ and $p^s_{T}[k] = \frac{\exp(z_k^s/T)}{\sum_{j} \exp(z_j^s/T)}$. $T$ is a temperature hyper-parameter. Our defense objective function is
\begin{align}
    \mathcal{L}_{\rm defense} = \mathbb{E}_{(x,y)\sim D_{\rm val}} D_{\rm KL}[p^t_T\| p^s_T]
\end{align}
where $D_{\rm KL}[\cdot\| \cdot]$ is the KL divergence. 

\myparagraph{Qualitative Analysis.} To show the effectiveness of knowledge distillation, we visualize the penultimate feature representations of clean and backdoor images throughout the process of knowledge distillation, and plot in the top row of Figure~\ref{fig:cifar_tsne}. The compactness and separability of clean image clusters reflect the model's prediction ability on normal data. Also, if backdoor behaviors are cleansed, the backdoor images will fall into the corresponding clean clusters. In Fig.~\ref{fig:cifar_tsne:a}, we can see that the clean images form 10 clusters, indicating a high ACC of the teacher model. The backdoor images are distant to the clean images and form separate clusters. Hence the teacher model behaves abnormally on backdoor data. For the student model after adaptive layer-wise initialization in Fig.~\ref{fig:cifar_tsne:b}, clean images from the same class are still close to each other, showing that some benign knowledge are preserved after initialization. This provides a good starting point for the following knowledge distillation. Figures~\ref{fig:cifar_tsne:c}-\ref{fig:cifar_tsne:e} show the results after training for some epochs. The normal behaviors are gradually transferred to the student model. With this, clean images form tighter clusters and are better separated. Backdoor images turn to overlap with the clean images with the same class labels, showing that the backdoor behaviors are successfully cleansed.

\subsection{Adaptive Layer-wise Initialization}
\label{sec:method:initialization}
It is generally believed that backdoor behavior is embodied through ``bad'' neurons. 
By random weight initialization and knowledge distillation on clean samples, we expect such neurons will be naturally cleansed. Previous observations~\cite{lyu2022study} reveal that these ``bad'' neurons can be distributed differently at different layers, and the distribution is architecture- and dataset-dependent.
In order to (1) break connection between triggers and target label and (2) preserve more normal knowledge simultaneously, we propose an adaptive layer-wise initialization strategy to initialize the student model. 

Assuming the suspicious teacher model has $L$ layers, the weights can be represented as $W^t = \{W_l^t | 1 \leq l \leq L\}$ where $W_l^t \in \mathbb{R}^{C_{\rm out}\times C_{\rm in}\times K\times K}$ for a convolution layer and $W_l^t \in \mathbb{R}^{C_{\rm out}\times C_{\rm in}}$ for a linear layer. We also have another random initialized student model, whose architecture is same as teacher model. Similarly, the weights of random initialized student model can be represented as $W^s = \{W_l^s | 1 \leq l \leq L\}$ where $W_l^s \in \mathbb{R}^{C_{\rm out}\times C_{\rm in}\times K\times K}$ for a convolution layer and $W_l^s \in \mathbb{R}^{C_{\rm out}\times C_{\rm in}}$ for a linear layer. Here, we consider a tuned hyperparameter $\delta_l$ for $l$-th layer. Then the initialization mask is defined as 
\begin{displaymath}
M = \{m_l|1\leq l \leq L, m_l \in \{0, 1\}^{{\rm shape}(W_l^s)}, \sum m_l = \delta_l |m_l|\} 
\end{displaymath}
where $|m_l|$ is the size of initializing mask. Then, initialized student model $W^{s*}$ can be formulated as follows:
\begin{align}
    ALI(W^{s*}, \delta) = \bigcup_{i=1}^L ((1 - m_l)\odot W_l^t + m_l\odot W_l^s)
\end{align}
where $\delta = \{\delta_l| 1\leq l \leq L\}$ is the ratio of random initializing weights per layer.

\myparagraph{Qualitative Analysis.} Similar to previous analysis in Sec.~\ref{sec:method:initialization}, we study the effects of adaptive layer-wise initialization for the student model through visualizing clean and backdoor sample features. The comparison strategies include uniform initialization that uses a same random initialization ratio for every layer, and single-layer initialization. To match our adaptive layer-wise initialization, we choose a specific ratio for the uniform initialization so that the total number of randomized weights equals in the two strategies. The same ratio is used for single-layer initialization. 

Comparing Figure~\ref{fig:cifar_tsne:f} with Figure~\ref{fig:cifar_tsne:b}, we can find that uniform initialization breaks the connection between trigger and target label. However, the benign information is also discarded as all clean images clutter together in the figure. From Fig.~\ref{fig:cifar_tsne:g}-\ref{fig:cifar_tsne:j}, When randomly initialize shallow layers like 1st or 2nd layer, the connection between trigger and target label is not broken while the clustering structure of clean images are destroyed. When randomly initialize deep layers like 3rd or 4th, the clean information can be preserved. The backdoor information is also partially eliminated in Fig.~\ref{fig:cifar_tsne:i}, where backdoor images become more dispersed. Therefore, to make a balance between preserving clean information and discarding backdoor information, it is better to use higher random initialization ratios for deeper layers and smaller ratios for shallow ones. This justifies the motivation of our adaptive layer-wise initialization.

\begin{algorithm}[t]
	\renewcommand{\algorithmicrequire}{\textbf{Input:}}
	\renewcommand{\algorithmicensure}{\textbf{Output:}}
	\caption{Backdoor Cleansing with Unlabeled Data} 
	\label{alg: distillation} 
	\begin{algorithmic}[1]
		\Require Backdoor model $f^t$ with weights $W^t$, random initialized student model $f^s$ with weights $W^s$, adaptive ratios $\delta$, unlabeled clean data $\mathcal{D}_{\rm defense}$, training epochs $E$, iterations per epoch $I$ and temperature $T$.
		\Ensure Clean model $f$
		\For {$l=0$ \textbf{to} $|W^t|$}
		\State Sample $R_l ^{{\rm shape}(W^t_l)} \thicksim {\rm Uniform}(0, 1)$
		\State Obtain boolean weight mask $m_l = \mathbb{I}[R_l < \delta_l]$
		\State $W^{s}_l = (1 - m_l)\odot W_l^t + m_l\odot W_l^s$
		\EndFor
		
		\For{$e=0$ \textbf{to} $E$}	
		\For{$i=0$ \textbf{to} $I$}				
		\State Sample mini-batches $\mathcal{B}_{\rm val}$ from $\mathcal{D}_{\rm defense}$
		\State Obtain temperature scaled probability $p^t_{T}$ from $f^t$, and  $p^s_{T}$ from $f^s$
		\State Update student model weights $W^s$ with $\mathcal{L}_{\rm defense} = D_{\rm KL}[p^t_{T}\| p^s_{T}]$
		\EndFor
		\State $f \leftarrow f^s$
		\EndFor	
	\end{algorithmic}
\end{algorithm}

\renewcommand{\tabcolsep}{0.15cm}
\setlength{\aboverulesep}{1pt}
\setlength{\belowrulesep}{1pt}
\begin{table*}[]
    \centering
    \scriptsize
    \begin{tabular}{p{0.9cm}<{\centering}||p{0.5cm}<{\centering}p{0.6cm}<{\centering}||p{0.5cm}<{\centering}p{0.6cm}<{\centering}|p{0.5cm}<{\centering}p{0.6cm}<{\centering}|p{0.5cm}<{\centering}p{0.6cm}<{\centering}|p{0.5cm}<{\centering}p{0.6cm}<{\centering}|p{0.5cm}<{\centering}p{0.6cm}<{\centering}|p{0.5cm}<{\centering}p{0.6cm}<{\centering}||p{0.5cm}<{\centering}p{0.6cm}<{\centering}|p{0.5cm}<{\centering}p{0.6cm}<{\centering}}
        \toprule
        \multirow{4}{*}{\tabincell{c}{Backdoor\\Attacks}} & \multicolumn{2}{c||}{\multirow{2}{*}{Original}} & \multicolumn{12}{c||}{In-distribution Labeled} & \multicolumn{4}{c}{In-distribution Unlabeled} \\
        \cmidrule{4-19}
         &  \multicolumn{2}{c||}{} & \multicolumn{2}{c|}{Finetuning} & \multicolumn{2}{c|}{Fine-pruning} & \multicolumn{2}{c|}{MCR (t=0.3)} & \multicolumn{2}{c|}{ANP} & \multicolumn{2}{c|}{NAD} & \multicolumn{2}{c||}{I-BAU} & \multicolumn{2}{c}{Ours}  & \multicolumn{2}{c}{Ours$^*$} \\
        \cmidrule{2-19}
         & ASR & ACC & ASR & ACC & ASR & ACC & ASR & ACC & ASR & ACC & ASR & ACC & ASR & ACC & ASR & ACC & ASR & ACC\\
        \midrule
       Badnets & 99.93 & 92.76 & 9.70 & 92.55 & 32.36 & 92.57 & 1.68 & 86.41 & 2.56 & 88.58 & 4.67 & 92.35 & 10.16 & 91.98 & 3.00 & 92.15 & 3.00 & 92.75\\
        Blended & 100.00 & 94.48 & 5.20 & 93.44 & 20.62 & 93.70 & 6.39 & 87.51 & 0.87 & 92.85 & 5.06 & 93.24 & 6.19 & 92.71 & 4.90 & 93.16 & 5.10 & 93.65\\
        IAB & 91.35 & 87.46 & 9.46 & 86.91 & 2.45 & 86.89 & 1.35 & 85.29 & 0.60 & 85.37 & 2.17 & 86.76 & 7.57 & 85.64 & 1.96 & 86.42 & 1.90 & 86.85\\
        LC & 99.55 & 94.51 & 97.14 & 93.49 & 60.23 & 93.88 & 5.33 & 88.18 & 4.62 & 91.30 & 52.74 & 93.38 & 21.41 & 92.72 & 1.81 & 93.17 & 1.40 & 93.66\\
        SIG & 95.09 & 93.71 & 5.41 & 93.16 & 5.66 & 93.55 & 2.33 & 87.69 & 0.41 & 92.09 & 1.88 & 92.95 & 15.76 & 92.45 & 0.91 & 92.58 & 1.18 & 93.14\\
        WaNet & 97.15 & 93.53 & 0.98 & 92.34 & 13.99 & 92.92 & 1.14 & 91.08 & 0.31 & 90.61 & 1.03 & 92.22 & 1.73 & 91.62 & 9.86 & 92.05 & 16.67 & 92.64\\
        \midrule
        Mean & 97.18 & 92.74 & 21.31 & 91.98 & 22.55 & 92.25 & 3.04 & 87.69 & 1.56 & 90.14 & 11.26 & 91.82 & 10.47 & 91.19 & 3.74 & 91.59 & 4.87 & 92.11\\
        \midrule
        Drop $\downarrow$ & -- & -- & 75.86 & 0.76 & 74.63 & 0.49 & 94.14 & 5.05 & 95.62 & 2.61 & 85.92 & 0.92 & 86.71 & 1.56 & 93.44 & 1.15 & 92.30 & 0.63\\
        \bottomrule
    \end{tabular}
    \caption{Defense results on backdoor models trained on CIFAR10. ($^*$Using double unlabeled data.)}
    \label{tab:cifar10-result}
\end{table*}

\renewcommand{\tabcolsep}{0.15cm}
\setlength{\aboverulesep}{1pt}
\setlength{\belowrulesep}{1pt}
\begin{table*}[]
    \centering
    \scriptsize
    \begin{tabular}{p{0.9cm}<{\centering}||p{0.5cm}<{\centering}p{0.6cm}<{\centering}||p{0.5cm}<{\centering}p{0.6cm}<{\centering}|p{0.5cm}<{\centering}p{0.6cm}<{\centering}|p{0.5cm}<{\centering}p{0.6cm}<{\centering}|p{0.5cm}<{\centering}p{0.6cm}<{\centering}|p{0.5cm}<{\centering}p{0.6cm}<{\centering}|p{0.5cm}<{\centering}p{0.6cm}<{\centering}||p{0.5cm}<{\centering}p{0.6cm}<{\centering}|p{0.5cm}<{\centering}p{0.6cm}<{\centering}}
        \toprule
        \multirow{4}{*}{\tabincell{c}{Backdoor\\Attacks}} & \multicolumn{2}{c||}{\multirow{2}{*}{Original}} & \multicolumn{12}{c||}{In-distribution Labeled} & \multicolumn{4}{c}{In-distribution Unlabeled} \\
        \cmidrule{4-19}
         &  \multicolumn{2}{c||}{} & \multicolumn{2}{c|}{Finetuning} & \multicolumn{2}{c|}{Fine-pruning} & \multicolumn{2}{c|}{MCR (t=0.3)} & \multicolumn{2}{c|}{ANP} & \multicolumn{2}{c|}{NAD} & \multicolumn{2}{c||}{I-BAU} & \multicolumn{2}{c}{Ours}  & \multicolumn{2}{c}{Ours$^*$} \\
        \cmidrule{2-19}
         & ASR & ACC & ASR & ACC & ASR & ACC & ASR & ACC & ASR & ACC & ASR & ACC & ASR & ACC & ASR & ACC  & ASR & ACC \\
        \midrule
       Badnets & 100.0 & 97.22 & 99.99 & 99.80 & 97.71 & 99.54 & 61.26 & 99.51 & 19.00 & 89.47 & 9.22 & 99.79 & 0.00 & 99.66 & 0.02 & 96.75 & 0.00 & 97.88\\
        Blended & 100.0 & 98.89 & 5.45 & 99.81 & 5.80 & 99.73 & 1.69 & 99.71 & 0.14 & 98.47 & 0.38 & 99.84 & 1.00 & 99.77 & 0.50 & 97.32 & 0.37 & 98.90\\
        IAB & 98.74 & 98.01 & 58.91 & 99.79 & 2.23 & 99.80 & 3.94 & 99.79 & 0.08 & 96.39 & 46.94 & 99.88 & 0.02 & 99.80 & 0.15 & 96.91 & 0.07 & 98.07\\
        LC & 94.74 & 95.75 & 67.68 & 99.74 & 96.37 & 99.59 & 3.07 & 99.50 & 0.11 & 94.15 & 37.82 & 99.72 & 0.03 & 99.71 & 0.86 & 96.64 & 0.81 & 96.60\\
        SIG & 97.80 & 98.87 & 96.59 & 99.84 & 99.06 & 99.80 & 93.06 & 99.74 & 78.43 & 98.22 & 96.64 & 99.86 & 30.54 & 99.77 & 1.59 & 97.09 & 6.31 & 98.71\\
        WaNet & 93.58 & 98.69 & 0.61 & 99.84 & 9.73 & 99.84 & 0.12 & 99.85 & 0.00 & 98.36 & 0.01 & 99.88 & 0.01 & 99.81 & 0.11 & 97.59 & 0.02 & 98.80\\
        \midrule
        Mean & 97.48 & 97.91 & 54.87 & 99.81 & 51.82 & 99.72 & 27.19 & 99.68 & 16.29 & 95.84 & 31.83 & 99.83 & 5.27 & 99.75 & 0.54 & 97.05 & 1.26 & 98.16\\
        \midrule
        Drop $\downarrow$ & -- & -- & 42.60 & -1.90 & 45.66 & -1.81 & 70.29 & -1.78 & 81.18 & 2.06 & 65.64 & -1.92 & 92.21 & -1.85 & 96.94 & 0.86 & 96.21 & -0.25\\
        \bottomrule
    \end{tabular}
    \vspace{-2mm}
    \caption{Defense results on backdoor models trained on GTSRB. ($^*$Using double unlabeled data.)}
    \label{tab:gtsrb-result}
    \vspace{-1mm}
\end{table*}

\section{Experiments}
\label{sec:experiments}
\subsection{Experiment settings}
\vspace{-2mm}\myparagraph{Datasets and Architecture.}
We conduct all backdoor models on two datasets include CIFAR10~\cite{krizhevsky2009cifar10} and GTSRB~\cite{stallkamp2012gtsrb}. For CIFAR10 and GTSRB, we split their original test datasets into defense dataset and test dataset. The total size of each defense dataset is 5000. Tiny-ImageNet~\cite{wu2017tiny} is used as the out-of-distribution dataset. We also construct another out-of-distribution dataset ``Tiny-ImageNet++" from ImageNet~\cite{deng2009imagenet}. Tiny-ImageNet++ contains 20,000 images distributed evenly in 1000 classes. Its image resolution is the same as Tiny-ImageNet. ResNet-18~\cite{he2016resnet} is adopted as the model architecture. From shallow to deep, ResNet-18 includes 1 convolution layer, 8 basic blocks and 1 FC layer. Except for FC layer, the more shallow the layer is, the less the weights are. The ratios of first convolution layer and FC layer are set 0.01 and 0.1. The ratios of eight basic blocks are 0.01, 0.01, 0.03, 0.03, 0.09, 0.09, 0.27 and 0.27.

\myparagraph{Backdoor attacks setting.}
We evaluate all defenses on six representative backdoor attacks including Badnets~\cite{gu2019badnets}, Blended attack~\cite{chen2017blended}, Label-consistent backdoor attack (LC)~\cite{turner2019lc}, Sinusoidal signal backdoor attack (SIG)~\cite{barni2019sig}, Input-aware dynamic backdoor attack (IAB)~\cite{nguyen2020IAB} and WaNet~\cite{nguyen2021wanet}. LC and SIG represent two classic clean-label backdoor attacks. Badnets, Blended, IAB and WaNet are representatives of label-poisoned backdoor attacks. Specifically, Badnets is a patch-based visible backdoor attack. Blended is a noise-based invisible attack. IAB is a dynamic backdoor attack. WaNet is an image-transformation-based invisible attack. For a fair comparison, the poison ratio for label-poisoned attacks is set as 0.1. For label-poisoned attacks, we poison 80\% samples of target label. The \textit{all-to-one} strategy is adopted for all backdoor attacks.

\myparagraph{Backdoor defense setting.}
We compare our method with six state-of-the-art defense methods including standard finetuning, Fine-pruning~\cite{liu2018fine-pruning}, Mode Connectivity Repair (MCR)~\cite{zhao2020MCR}, Adversarial Neuron Pruning (ANP)~\cite{wu2021ANP}, Neural Attention Distillation (NAD)~\cite{li2021NAD} and Implicit Backdoor Adversarial Unlearning (I-BAU)~\cite{zeng2021i-bau}. 

For each attack, we train 14 backdoor models with different target labels and random seeds. We conduct all defenses on 14 models and the average is the final results. For fair comparison, we train 100 epochs for all defense methods. We set the batch size as 256 and optimize our framework using Stochastic Gradient Descent (SGD) with a momentum of 0.9, and a weight decay of 0.0005. The adopted data augmentation techniques include random crop and random horizontal flipping. For MCR, we get a benign model by finetuning the original backdoor model with 10 epochs.

\subsection{Comparison with other defense methods}

\vspace{-2mm}\myparagraph{Results using unlabeled in-distribution data.}
We compare with six state-of-the-art defenses with regard to ACC and ASR. Other six defenses use labeled clean samples, while our framework uses unlabeled samples. We assume that all defenses can access 2500 clean samples. For our method, we also present results using 5000 unlabeled samples in the last two columns. Results on CIFAR10~\cite{krizhevsky2009cifar10} and GTSRB~\cite{stallkamp2012gtsrb} are shown in Table \ref{tab:cifar10-result} and Table \ref{tab:gtsrb-result}, separately. Despite that our framework is trained without using ground-truth labels, its performance is still comparative with other methods that require labels. For CIFAR10, due to the usage of labels, existing works get the highest ACC of 92.25\%. However, these works can not decrease ASRs largely while keep high ACC. Our method reduces ASR to 3.74\% with negligible ACC reduction of 1.15\%. For GTSRB, since ground-truth labels are utilized, ACCs increase slightly in five of six defenses. However, our framework obtains a robust model by reducing average ASR to less than 1\%, which is better than other label-based methods. Meanwhile, the ACC reduction of our framework is only 0.86\%. With 5000 unlabeled data, our ACC increases 0.25\%.

For both datasets, ANP succeeds in dropping ASR of most attacks, but at the expense of lower accuracies compared other methods. ANP aims to prune the bad neurons without re-training backdoor model. However, the backdoor neurons are difficult to distinguish from normal neurons in reality. Some neurons critical to ACC may be pruned by ANP, leading to degraded performances. Fine-pruning gets a low average drop over ASR since Fine-pruning simply prunes the dormant neurons in the last convolution layer. However, complex triggers activate neurons across different layers. Since a finetuning stage follows the pruning process, Fine-pruning has a high ACC. We find that finetuning, Fine-prunng and NAD perform badly on LC attack in reducing ASR. All of three defenses include a finetuning stage. Though NAD distillates attention map knowledge from teacher to student model, teacher model is obtained by finetuning backdoor model and student model is supervised by CrossEntropy loss. One possible reason is that the PGD perturbations used in LC hinder finetuning to associate normal images with target labels with limited clean samples. MCR introduces a curve model to find a path connection between two backdoor models. With limited data samples, MCR achieves low ACC compared other methods. In all six defenses, I-BAU perform well on both datasets. I-BAU adopt implicit hypergradient to solve minmax optimization, leading to strong generalizability of the robustness. Note that most defense methods can not defend SIG attack on GTSRB because we improve sinusoidal signal to inject backdoor successfully ($\Delta$ is set 60 in our experiments). This strong signal is not stealthy to GTSRB images, causing backdoor model learn strong abnormal behaviors and difficult to defend.

\renewcommand{\tabcolsep}{0.15cm}
\setlength{\aboverulesep}{1pt}
\setlength{\belowrulesep}{1pt}
\begin{table}[]
    \centering
    \scriptsize
    \begin{tabular}{p{0.9cm}<{\centering}|p{0.5cm}<{\centering}p{0.5cm}<{\centering}|p{0.4cm}<{\centering}p{0.5cm}<{\centering}|p{0.4cm}<{\centering}p{0.5cm}<{\centering}|p{0.4cm}<{\centering}p{0.5cm}<{\centering}}
        \toprule
        \multirow{4}{*}{\tabincell{c}{Backdoor\\Attacks}} & \multicolumn{2}{c|}{In distribution} & \multicolumn{6}{c}{Out-of-distribution} \\
        \cmidrule{2-9}
         &  \multicolumn{2}{c|}{CIFAR10} &   \multicolumn{2}{c|}{GTSRB} & \multicolumn{2}{c}{Tiny-IN} & \multicolumn{2}{c}{Tiny-IN++} \\
        \cmidrule{2-9}
         & ASR & ACC & ASR & ACC & ASR & ACC & ASR & ACC\\
         \midrule
        Badnets & 3.00 & 92.15 & 11.30 & 81.19 & 4.47 & 91.24 & 3.03 & 92.44\\
        Blended & 4.90 & 93.16 & 6.68 & 82.39 & 61.75 & 92.88 & 11.87 & 93.66\\
        IAB & 1.96 & 86.42 & 1.62 & 81.91 & 1.52 & 86.00 & 1.52 & 86.76\\
        LC & 1.81 & 93.17 & 3.49 & 84.47 & 1.95 & 92.83 & 1.46 & 93.67\\
        SIG & 0.91 & 92.58 & 0.98 & 81.49 & 14.58 & 91.85 & 17.79 & 92.86\\
        WaNet & 9.86 & 92.05 & 83.89 & 84.11 & 7.62 & 91.58 & 22.60 & 92.52\\
        \midrule
        Mean & 3.74 & 91.59 & 17.99 & 82.59 & 15.32 & 91.06 & 9.71 & 91.99\\
        \bottomrule
    \end{tabular}
    \vspace{-2mm}
    \caption{Defense results on CIFAR10 using different unlabeled out-of-distribution data.}
    \vspace{-2mm}
    \label{tab:cifar10-result-out}
\end{table}

\renewcommand{\tabcolsep}{0.15cm}
\setlength{\aboverulesep}{1pt}
\setlength{\belowrulesep}{1pt}
\begin{table}[]
    \centering
    \scriptsize
    \begin{tabular}{p{0.9cm}<{\centering}|p{0.5cm}<{\centering}p{0.5cm}<{\centering}|p{0.4cm}<{\centering}p{0.5cm}<{\centering}|p{0.4cm}<{\centering}p{0.5cm}<{\centering}}
        \toprule
        \multirow{2}{*}{\tabincell{c}{Backdoor\\Attacks}} & \multicolumn{2}{c|}{Uniform} & \multicolumn{2}{c|}{\tabincell{c}{Adaptive \\ decreasing}} & \multicolumn{2}{c}{\tabincell{c}{Adaptive\\ increasing}}\\
        \cmidrule{2-7}
         & ASR & ACC & ASR & ACC & ASR & ACC\\
         \midrule
        Badnets & 4.88 & 92.08 & 2.38 & 86.75 & 3.00 & 92.15\\
        Blended & 4.54 & 93.01 & 3.32 & 88.33 & 4.90 & 93.16\\
        IAB & 1.72 & 86.25 & 2.68 & 81.51 & 1.96 & 86.42\\
        LC & 4.18 & 93.01 & 1.05 & 88.22 & 1.81 & 93.17\\
        SIG & 0.58 & 92.31 & 1.07 & 88.24 & 0.91 & 92.58\\
        WaNet & 7.35 & 91.69 & 2.17 & 84.76 & 9.86 & 92.05\\
        \midrule
        Mean & 3.87 & 91.39 & 2.11 & 86.30 & 3.74 & 91.59\\
        \bottomrule
    \end{tabular}
    \vspace{-2mm}
    \caption{Comparison of weights initialization strategies for student model on CIFAR10 (in-distribution).}
    \label{tab:cifar10-cmp-init}
    \vspace{-2mm}
\end{table}

\vspace{-2mm}\myparagraph{Results using unlabeled out-of-distribution data.}
We conduct experiments on CIFAR10 by using out-of-distribution unlabeled data. GTSRB, Tiny-ImageNet and Tiny-ImageNet++ are three out-of-distribution unlabeled datasets. Table \ref{tab:cifar10-result-out} reports the results. For GTSRB and Tiny-ImageNet, we random sample 2500 images from our constructed defense dataset.

Compared to in-distribution data, GTSRB reduces ASRs largely in five of six attacks while perform badly on WaNet. The possible reason is that simple GTSRB images e.g. circle or triangular signs, introduce warping-based backdoor behavior. Due to large domain gap between GTSRB and CIFAR10, GTSRB decreases average ACC about 10\%. With Tiny-ImageNet, our method can reduce ASRs largely especially for Badnets, IAB, LC and WaNet, with negligible ACC cost. However, Tiny-ImageNet can not reduce ASR successfully on Blended Attack. Meanwhile, Tiny-ImageNet++ reduces ASR to 11.87\% on LC. Blended trigger is a random noise. Removing random noise trigger needs more out-of-distribution natural clean images.
Due to the large size and diversity, Tiny-ImageNet++ performs better than GTSRB and Tiny-ImageNet. Tiny-ImageNet++ reduces average ASR to less than 10\%, while other two datasets reduce ASRs to more than 15\%. Tiny-ImageNet++ can also keep ACC high after defense. 


\begin{figure*}[!t]
	\centering	
	\begin{subfigure}[]{0.33\textwidth}
		\includegraphics[width=\textwidth]{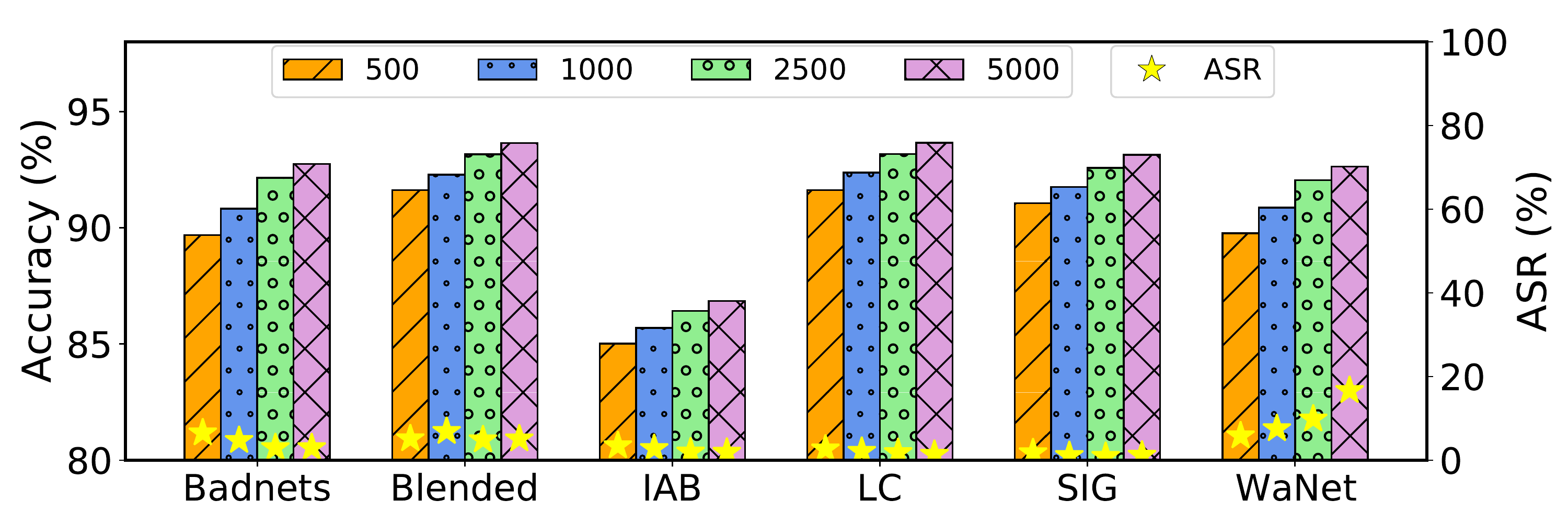}
		\caption{In-distribution CIFAR10}	
	\end{subfigure}
	\hfill
	\begin{subfigure}[]{0.33\textwidth}
		\includegraphics[width=\textwidth]{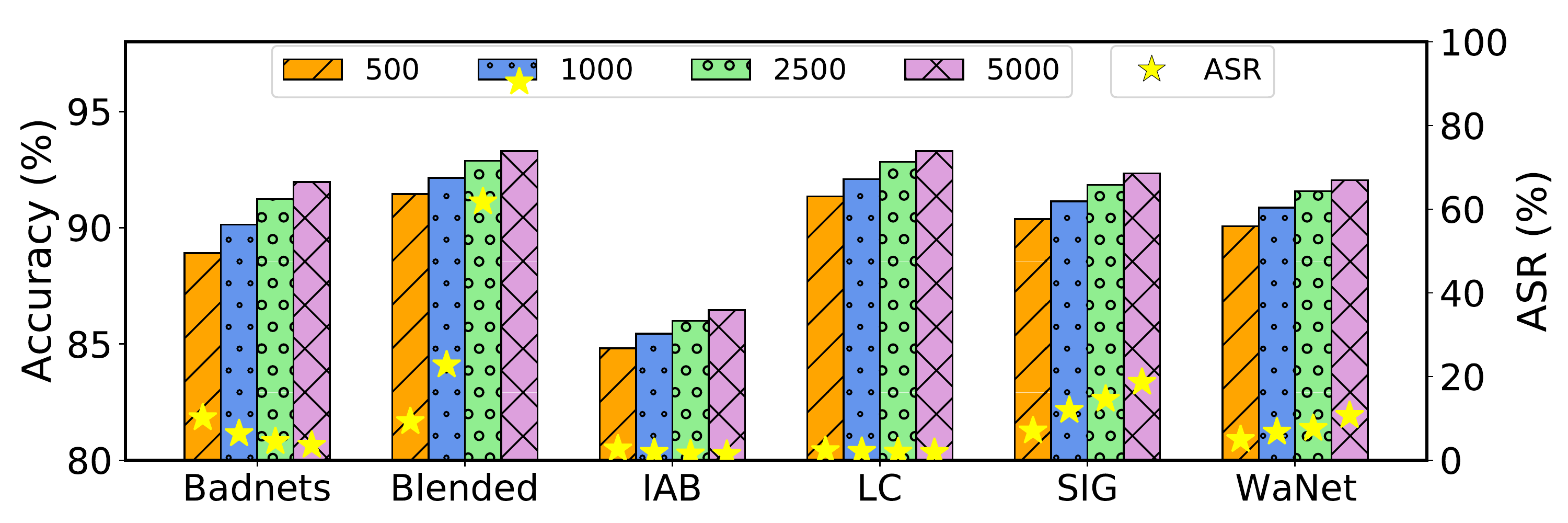}
		\caption{Out-of-distribution Tiny-ImageNet}	
	\end{subfigure}
	\hfill
	\begin{subfigure}[]{0.33\textwidth}
		\includegraphics[width=\textwidth]{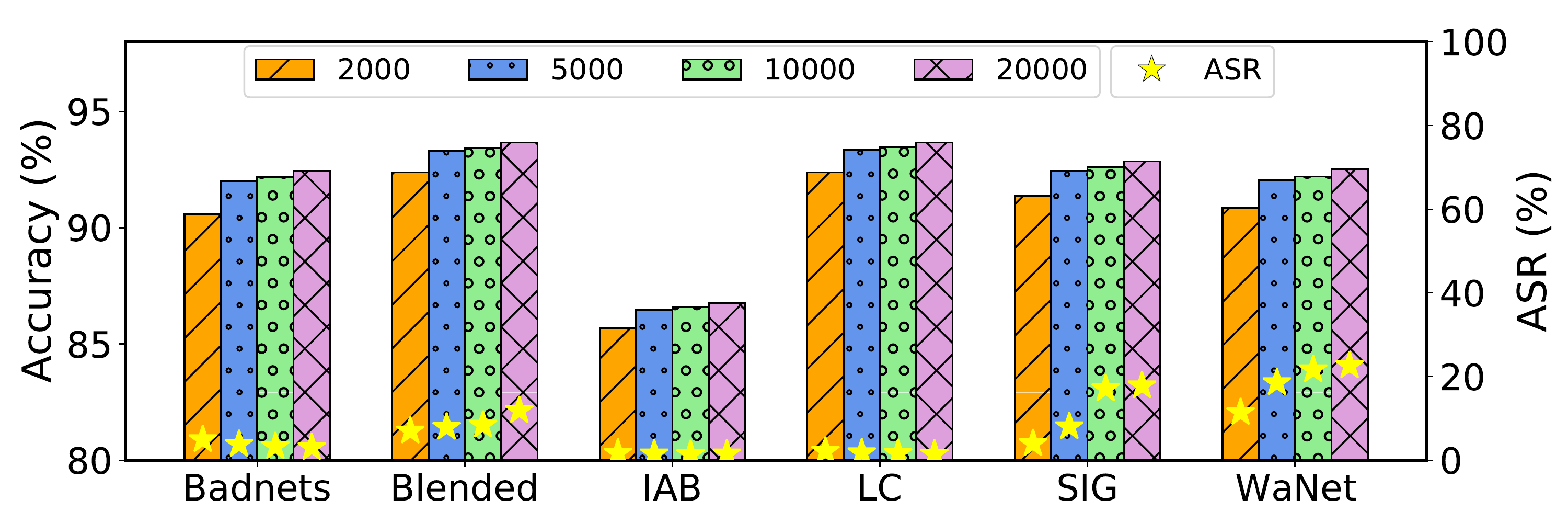}
		\caption{Out-of-distribution Tiny-ImageNet++}	
	\end{subfigure}
	\vspace{-2mm}
	\caption{Defense results on CIFAR10 using different numbers of unlabeled samples.}
	\label{fig:cifar_size}
	\vspace{-3mm}
\end{figure*}

\subsection{Analysis}
\vspace{-2mm}\myparagraph{Size of unlabeled samples.}
We use CIFAR10 to analyze influence of the size of unlabeled samples. Figure \ref{fig:cifar_size} (a--c) show the results using in-distribution CIFAR10, out-of-distribution Tiny-ImageNet and Tiny-ImageNet++. For three datasets, we randomly sample 500, 1000, 2500, 5000 images, separately. As the number of samples increases, ACCs increase and ASRs decrease for most cases. However, with the number of unlabeled Tiny-ImageNet and Tiny-ImageNet++ data increasing, ASRs raise up on Blended, SIG and WaNet attacks. Blended attack injects backdoor by blending clean images and random noise. The trigger of SIG is a sinusoidal signal. WaNet applies elastic warping to design triggers. All three triggers are stealthy and cause slight change to images. Some images in Tiny-ImageNet++ are downloaded from the internet and might include light noise similar to the three triggers. Therefore, using more out-of-distribution unlabeled images from Tiny-ImageNet or Tiny-ImageNet++ might cause ASRs increasing for the three attacks.

\vspace{-5mm}\paragraph{Adaptive layer-wise initialization.}
We analyze the effectiveness of different adaptive layer-wise initialization strategies by conducting experiments on CIFAR10. Three strategies are designed including random initialize weights of student model with uniform ratio, increasing ratio and decreasing ratio. For fair comparison, the overall ratio of random initialization keeps around $0.2$ for three strategies. The results are presented in Table \ref{tab:cifar10-cmp-init}. All of three strategies can reduce ASRs to less than 5\%. However, adaptive decreasing layer-wise initialization performs bad on ACCs. The reason is that random initializing two many weights in low layers causes student model dropping too much information related to low-level features. It is difficult to recover effectively only by aligning two probability distributions between student and teacher models. Compared to uniform initializing strategy, adaptive increasing layer-wise initialization obtains lowest ASR and highest ACC. 

\vspace{-5mm}\paragraph{Effectiveness of knowledge distillation.}
To evaluate the effectiveness of knowledge distillation, we compare the performances using soft labels and hard labels. Hard labels are class labels with the maximum probability of teacher model outputs. Soft labels are soft probability with temperature $T$ described in Section \ref{sec:method}. Cross-Entropy loss function is employed for hard labels setting. The experiments are conducted on CIFAR10 and out-of-distribution dataset is Tiny-ImagneNet. Table \ref{tab:cifar_hard} shows the results. It shows that hard and soft labels achieve comparative performance for in-distribution unlabeled data. The reason is that backdoor teacher model predicts high ACC for in-distribution images. Therefore, most hard labels are ground-truth labels. However, backdoor teacher model can not predict correct hard labels for out-of-distribution data. Some classes of out-of-distribution images even does not exist in the CIFAR10. Therefore, using soft labels is better than hard labels. Specifically, ASR of using soft labels is 1.21\% lower than ASR of using hard labels. ACC of using soft labels is 2.29\% higher than ACC of using hard labels.

\renewcommand{\tabcolsep}{0.15cm}
\setlength{\aboverulesep}{1pt}
\setlength{\belowrulesep}{1pt}
\begin{table}[]
    \centering
    \scriptsize
    \begin{tabular}{p{0.9cm}<{\centering}|p{0.5cm}<{\centering}p{0.5cm}<{\centering}|p{0.4cm}<{\centering}p{0.5cm}<{\centering}|p{0.4cm}<{\centering}p{0.5cm}<{\centering}|p{0.4cm}<{\centering}p{0.5cm}<{\centering}}
        \toprule
        \multirow{4}{*}{\tabincell{c}{Backdoor\\Attacks}} & \multicolumn{4}{c|}{In-distribution} & \multicolumn{4}{c}{Out-of-distribution (Tiny-IN)} \\
        \cmidrule{2-9}
         &  \multicolumn{2}{c|}{Soft} &   \multicolumn{2}{c|}{Hard} & \multicolumn{2}{c}{Soft} & \multicolumn{2}{c}{Hard} \\
        \cmidrule{2-9}
         & ASR & ACC & ASR & ACC & ASR & ACC & ASR & ACC\\
         \midrule
        Badnets & 3.00 & 92.15 & 3.37 & 91.16 & 4.47 & 91.24 & 5.55 & 88.74\\
        Blended & 4.90 & 93.16 & 5.14 & 92.09 & 61.75 & 92.88 & 69.48 & 90.71\\
        IAB & 1.96 & 86.42 & 1.64 & 85.31 & 1.52 & 86.00 & 2.05 & 83.85\\
        LC & 1.81 & 93.17 & 1.86 & 92.05 & 1.95 & 92.83 & 1.40 & 90.61\\
        SIG & 0.91 & 92.58 & 1.42 & 91.61 & 14.58 & 91.85 & 16.11 & 89.69\\
        WaNet & 9.86 & 92.05 & 3.16 & 90.75 & 7.62 & 91.58 & 4.59 & 89.03\\
        \midrule
        Mean & 3.74 & 91.59 & 2.77 & 90.50 & 15.32 & 91.06 & 16.53 & 88.77\\
        \bottomrule
    \end{tabular}
    \caption{Comparisons of using soft predictions and hard predictions of backdoor models for distillation on CIFAR10.}
    \vspace{-3.5mm}
    \label{tab:cifar_hard}
\end{table}

\begin{figure}[!t]
	\centering	
	\includegraphics[width=0.45\textwidth,height=.32\textwidth]{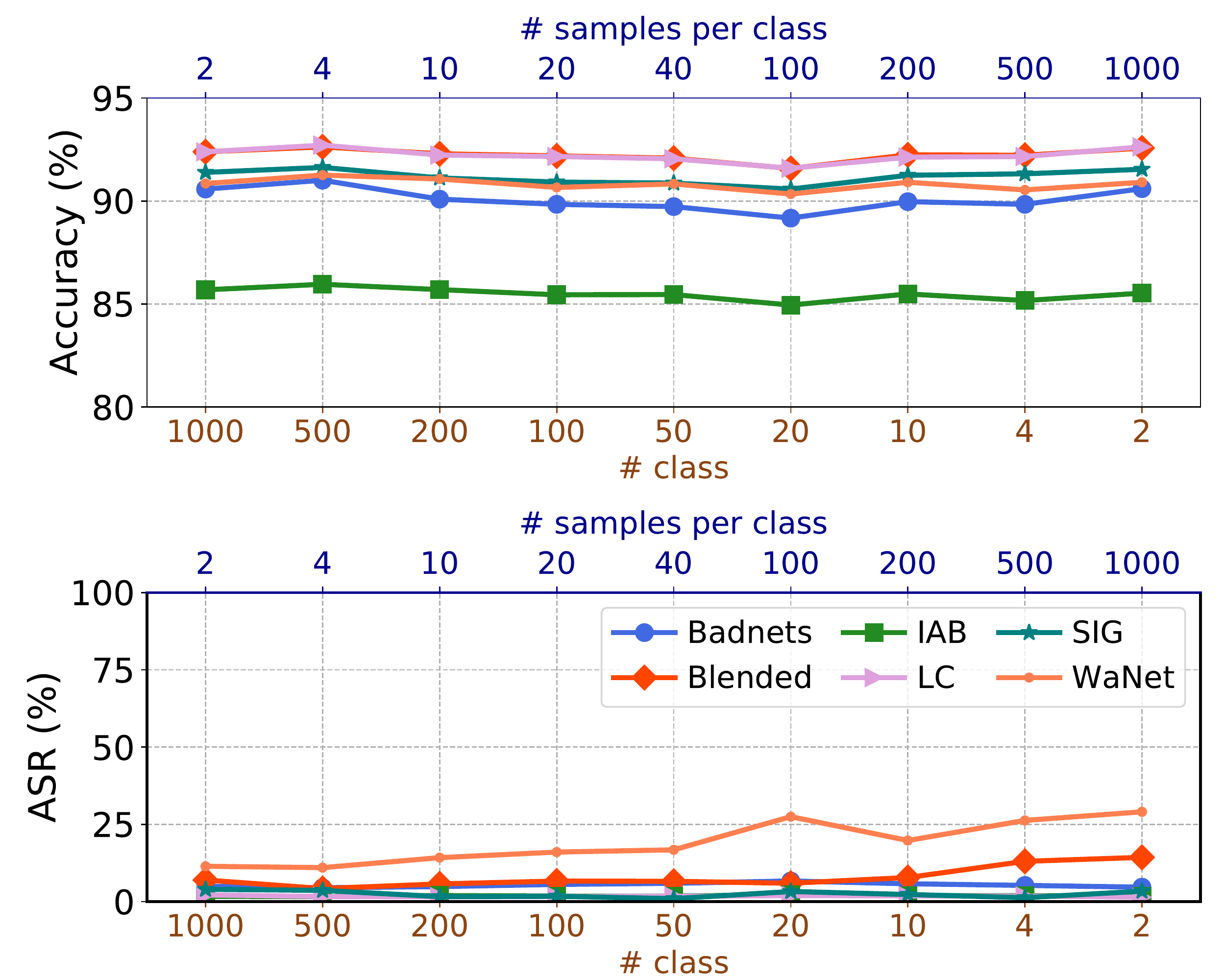}
	\vspace{-2mm}
	\caption{Defense results on CIFAR10 using Tiny-ImageNet++ created with different configurations. 
	}
	\label{fig:cifar_out_div}
	\vspace{-4mm}
\end{figure}

\vspace{-5mm}\paragraph{Diversity of out-of-distribution data.}
To study how diversity of out-of-distribution data influences defense performance, we create several versions of Tiny-ImangeNet++ with different configurations of (number of class, number of samples per class). The total number of unlabeled images is fixed to 2000. Then we apply them to cleanse backdoor models trained on CIFAR10. Figure~\ref{fig:cifar_out_div} plots the curves of ACC and ASR. ACCs are close for different configurations. However, as the unique number of classes in the training data increases, ASR has a tendency to decrease, showing that backdoor behaviors are more effectively eliminated. In principle, increasing the diversity of out-of-distribution unlabeled data is beneficial as more data modes are covered. It is more likely that data similar to the training distribution are included. Also, the student model can learn more general knowledge in making classification than specific ones.

\section{Conclusion}
\label{sec:conclusion}
In this paper, for the first time, we explore the possibility of using unlabeled data including in-distribution and out-of-distribution data to remove backdoor from a backdoor model. A knowledge distillation framework with a carefully designed adaptive layer-wise initialization strategy is proposed. We conduct experiments on two datasets including CIFAR10 and GTSRB against six representative backdoor attacks. Results show that our framework can successfully defend backdoor attacks with negligible clean accuracy decrease, compared with existing methods using labeled in-distribution data.

\vspace{-2mm}\paragraph{Acknowledgement.} This effort was partially supported by the Intelligence Advanced
Research Projects Agency (IARPA) and Army Research Office (ARO) under
Contract No.~W911NF20C0038, and by US National Science Foundation Grants (No.\ 2128187, No.\ 2128350 and No.\ 2006655). Any opinions, findings, and conclusions in this paper are those of the authors only and do not necessarily reflect the views of our sponsors.

{\small
\bibliographystyle{ieee_fullname}
\bibliography{egbib}
}

\appendix
\clearpage

\end{document}


\title{Backdoor Cleansing with Unlabeled Data ---- Supplementary Materials}
\author{Lu Pang, Tao Sun, Haibin Ling, Chao Chen\\
Stony Brook University\\
{\tt\small \{luppang,tao,hling\}@cs.stonybrook.edu, 
chao.chen.1@stonybrook.edu}}

\maketitle

\appendix

\section{Experimental Details on Backdoor Attacks}

We compare all backdoor defense methods against six backdoor attacks. For every dataset and every attack, we train 14 backdoor models in total using different target labels and random seeds. For CIFAR10~\cite{krizhevsky2009cifar10}, 5 models share Class 0 as the target label (with different random seeds), and the other 9 models use Class 1-9 as the target label, respectively. For GTSRB~\cite{stallkamp2012gtsrb}, we set target labels from the 10 major classes. Similarly, 5 models share the most major class, and the other 9 models use the rest 9 major classes.

\begin{figure}
    \centering
    \includegraphics[width=0.48\textwidth]{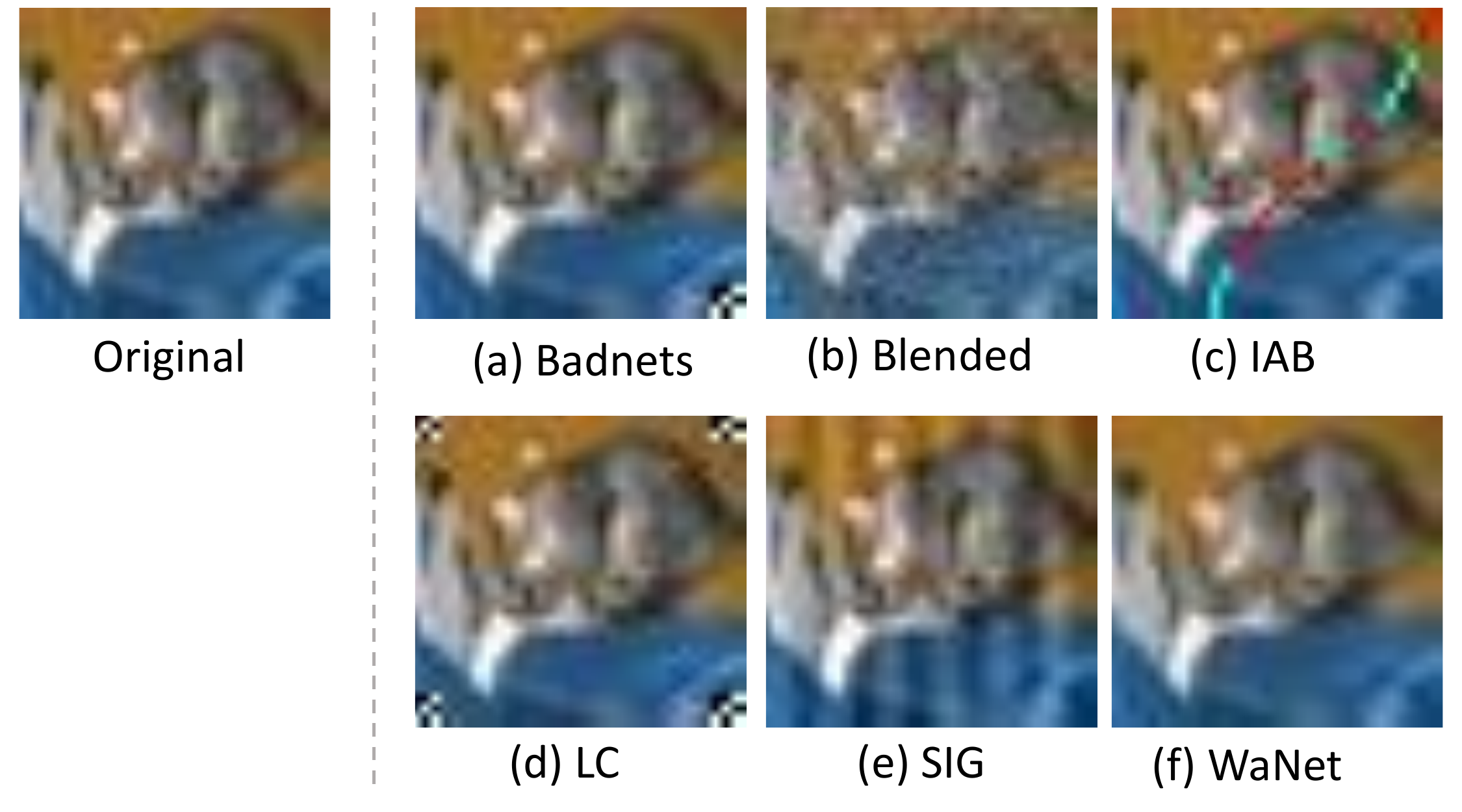}
    \caption{Poisoned images with triggers from 6 backdoor attacks.}
    \label{fig:trigger}
\end{figure}

We show poisoned images with triggers from 6 backdoor attacks in Figure~\ref{fig:trigger}. Other implementation details are as follows:

\myparagraph{Badnets~\cite{gu2019badnets}.} The trigger is a $3\times 3$ checkerboard located at the lower right corner of an image. We randomly choose $10\%$ training samples to attach triggers. The reported ACC and ASR are the average of $14$ models.

\myparagraph{Blended Attack~\cite{chen2017blended}.} We use random pattern as the trigger. Each pixel value in the random pattern is uniformly sampled over $[0, 256)$. Following the original paper, we attach trigger by using \textit{Blended Injection Strategy} and the corresponding blend ratio $\alpha$ is $0.2$. We randomly choose $10\%$ training samples to attach triggers.

\myparagraph{IAB~\cite{nguyen2020IAB}.} The trigger of IAB varies from sample to sample. Following original paper and open-source code\footnote{\url{https://github.com/VinAIResearch/input-aware-backdoor-attack-release}}, we train trigger generator and image classifier simultaneously. We adopt all-to-one strategy and the injection ratio is $0.1$.

\myparagraph{Label-Consistent Attack (LC)~\cite{turner2019lc}.} The trigger is four $3\times 3$ checkerboards at four corners of an image. We use projected gradient descent (PGD) to generate adversarial perturbations for misclassifying poisoned samples during the training process. The adversarial model is trained with bounded in $l_{\inf}$ norm and $\epsilon = 16$. We poison $80\%$ samples from the target label.

\myparagraph{SIG~\cite{barni2019sig}.} We employ sinusoidal backdoor signal with $f=6$ as the trigger. For CIFAR10, we follow the original paper to set $\Delta=20$. For GTARB, we set $\Delta=60$ to successfully attack models. Since both LC and SIG are clean-label backdoor attack, we also poison $80\%$ samples from the target label.

\myparagraph{WaNet~\cite{nguyen2021wanet}.} Following the original paper and open-source code\footnote{\url{https://github.com/VinAIResearch/Warping-based_Backdoor_Attack-release}}, we train a backdoor model with three modes. The backdoor probability $\rho_a$ and the noise probability $\rho_n$ are set as $0.1$ and $0.2$, respectively. The injection ratio is $0.1$.

For Badnets, Blended Attack, LC and SIG, we train $200$ epochs using Stochastic Gradient Descent (SGD) with a momentum of $0.9$ and a weight decay of $0.0005$. The initial learning rate is set as $0.1$. Following \textit{MultiStepLR} in PyTorch, the learning rate decays with \textit{milestones} of $[100, 150]$ and \textit{gamma} of $0.1$. For IAB and WaNet, we train backdoor models following released open-source codes.

\begin{figure*}[!t]
	\centering	
	\begin{subfigure}[]{0.195\textwidth}
		\includegraphics[width=\textwidth]{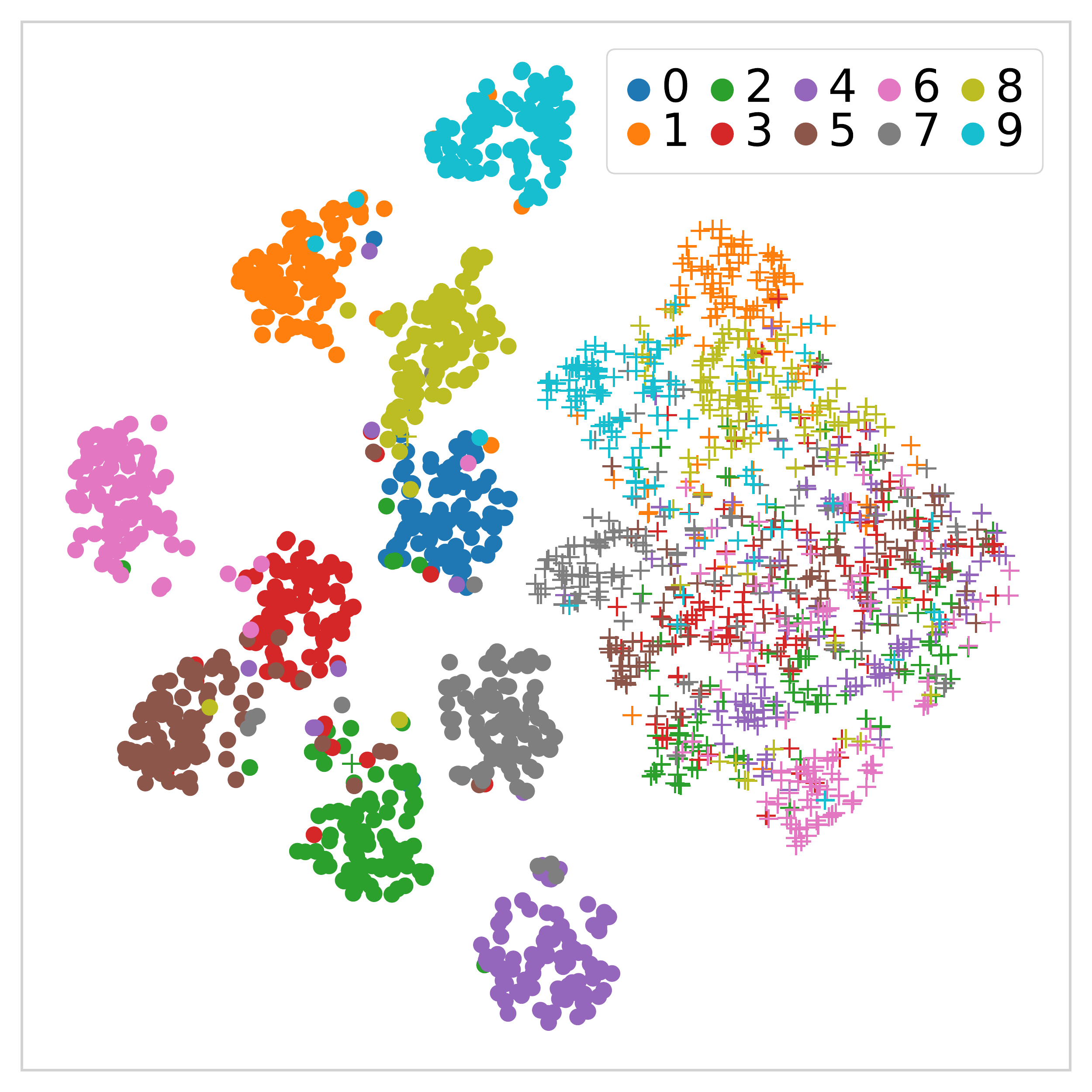}
		\caption{Teacher}	
		\label{fig:cifar_tsne:a}
	\end{subfigure}
	\begin{subfigure}[]{0.195\textwidth}
		\includegraphics[width=\textwidth]{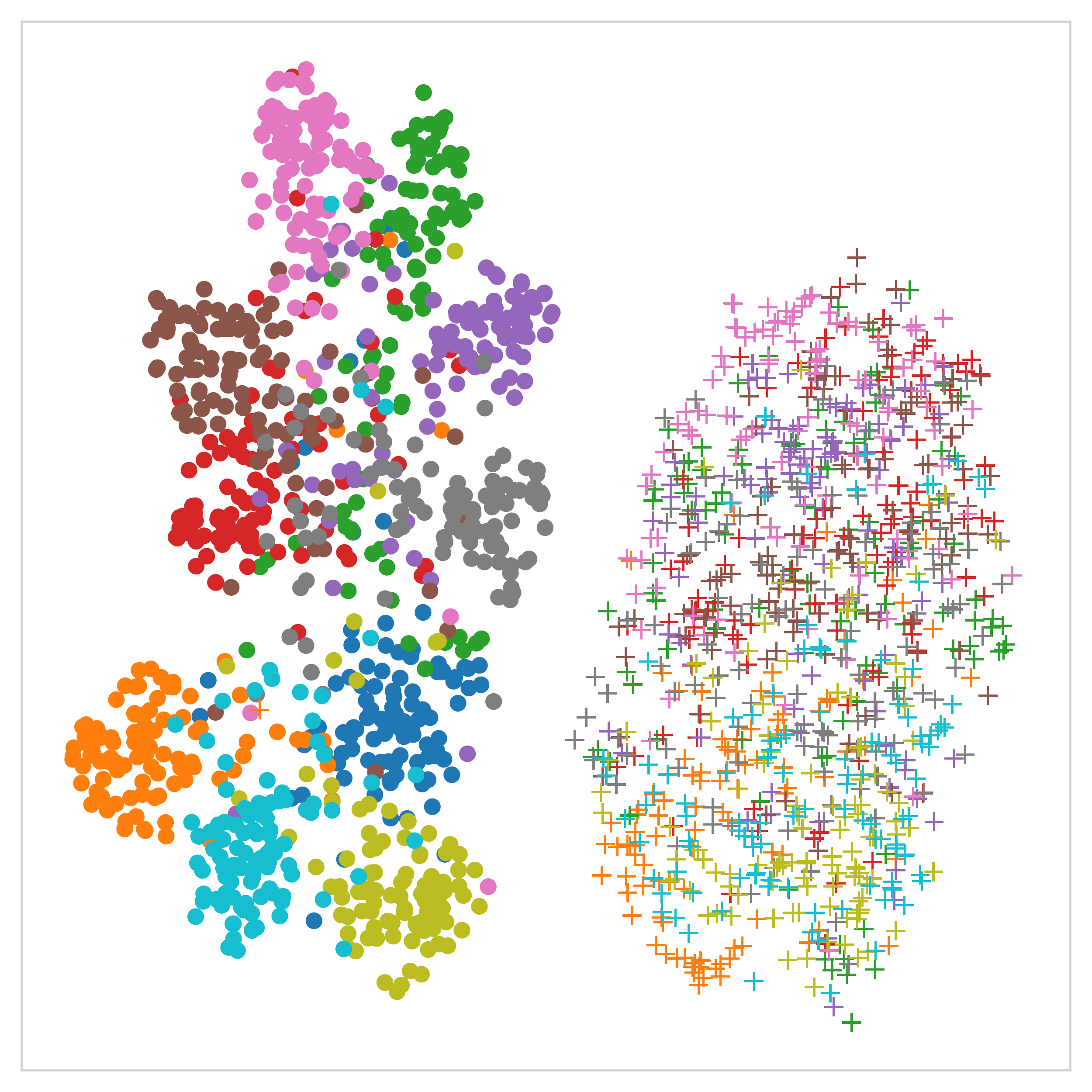}
		\caption{Stu (adaptive init, ep 0)}	
		\label{fig:cifar_tsne:b}
	\end{subfigure}
	\begin{subfigure}[]{0.195\textwidth}
		\includegraphics[width=\textwidth]{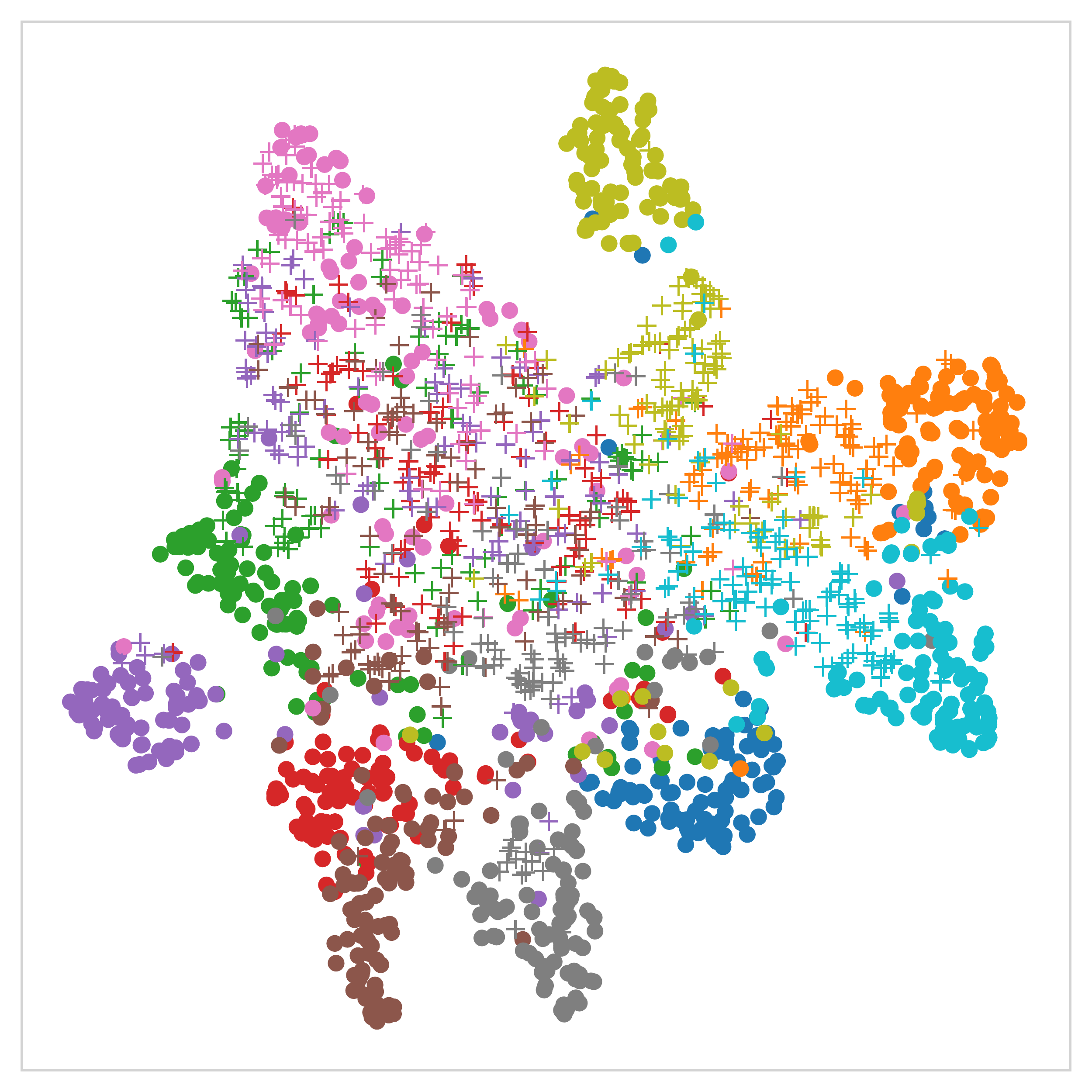}
		\caption{Stu (adaptive init, ep 1)}	
		\label{fig:cifar_tsne:c}
	\end{subfigure}
	\begin{subfigure}[]{0.195\textwidth}
		\includegraphics[width=\textwidth]{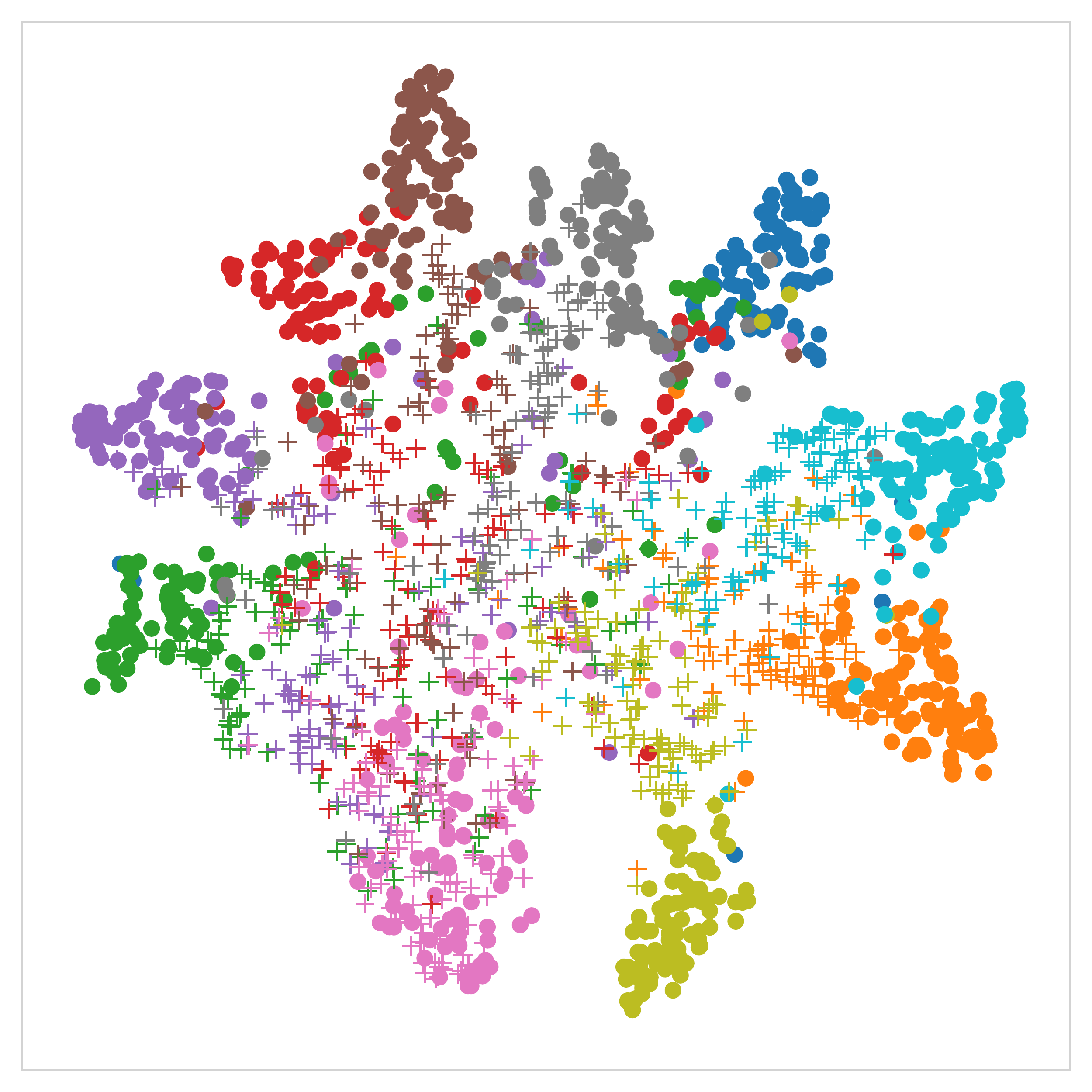}
		\caption{Stu (adaptive init, ep 2)}	
		\label{fig:cifar_tsne:d}
	\end{subfigure}
		\begin{subfigure}[]{0.195\textwidth}
		\includegraphics[width=\textwidth]{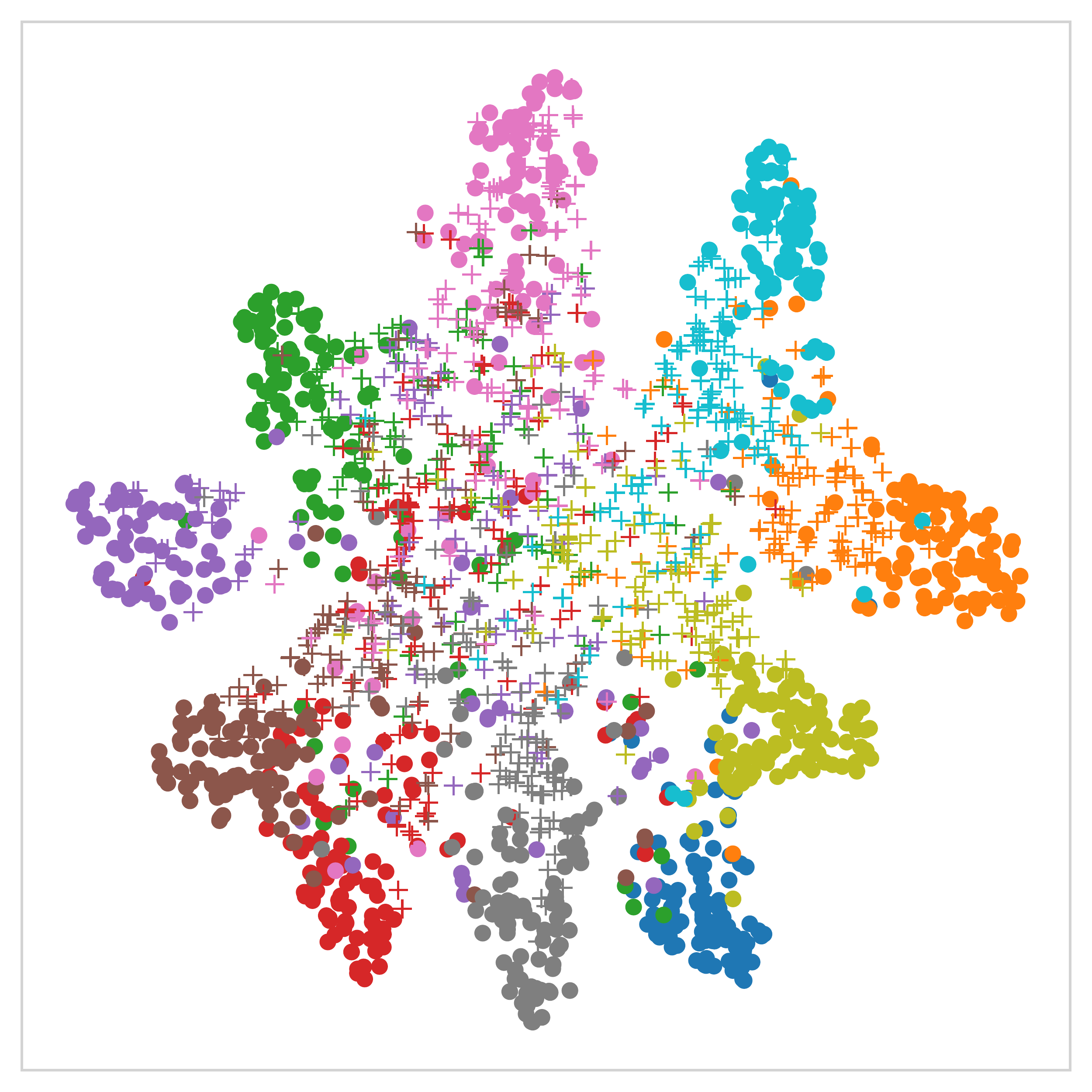}
		\caption{Stu (adaptive init, ep 11)}	
		\label{fig:cifar_tsne:e}
	\end{subfigure}
	
	\begin{subfigure}[]{0.195\textwidth}
	\includegraphics[width=\textwidth]{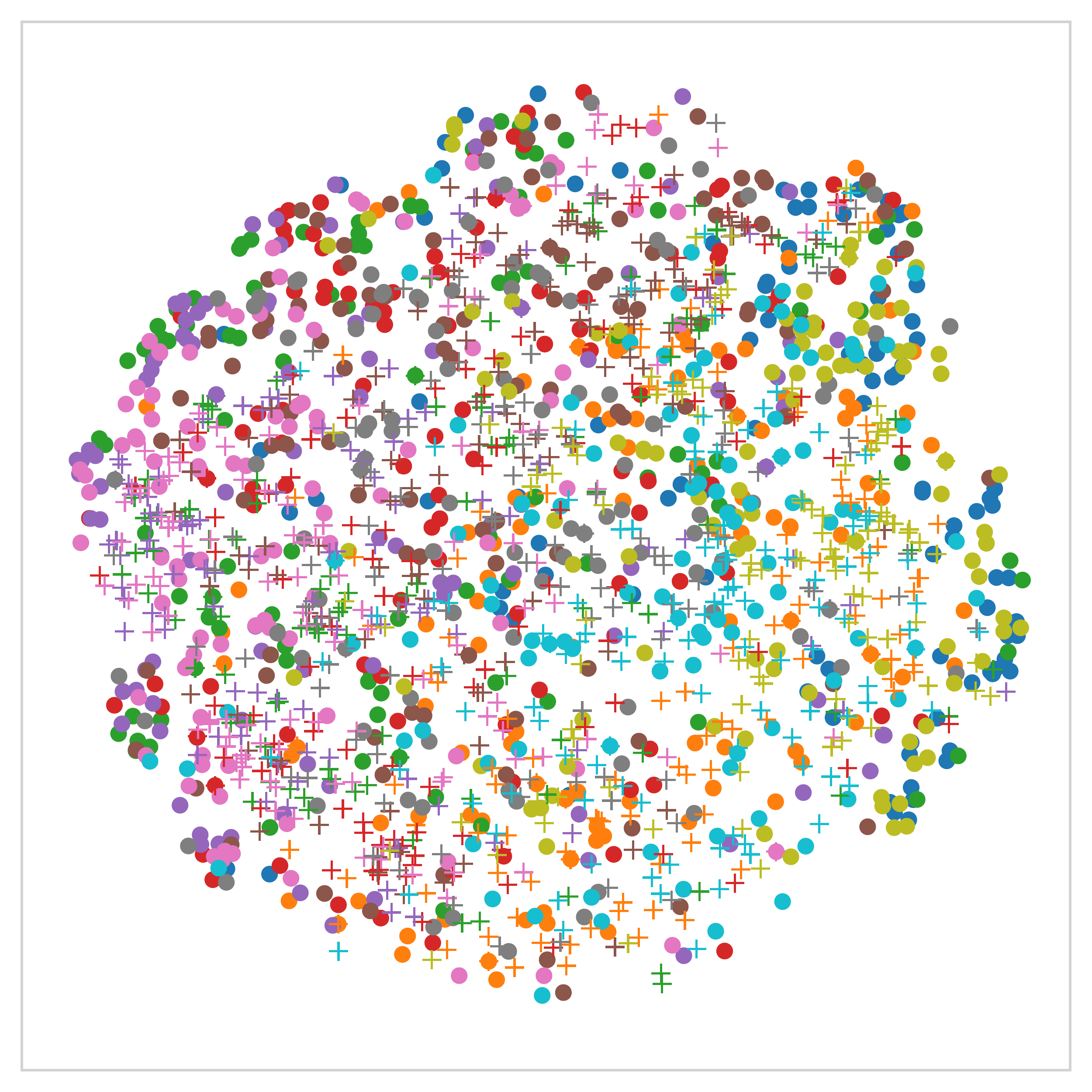}
	\caption{Stu (uniform init, ep 0)}
	\label{fig:cifar_tsne:f}
 	\end{subfigure}
	\begin{subfigure}[]{0.195\textwidth}
	\includegraphics[width=\textwidth]{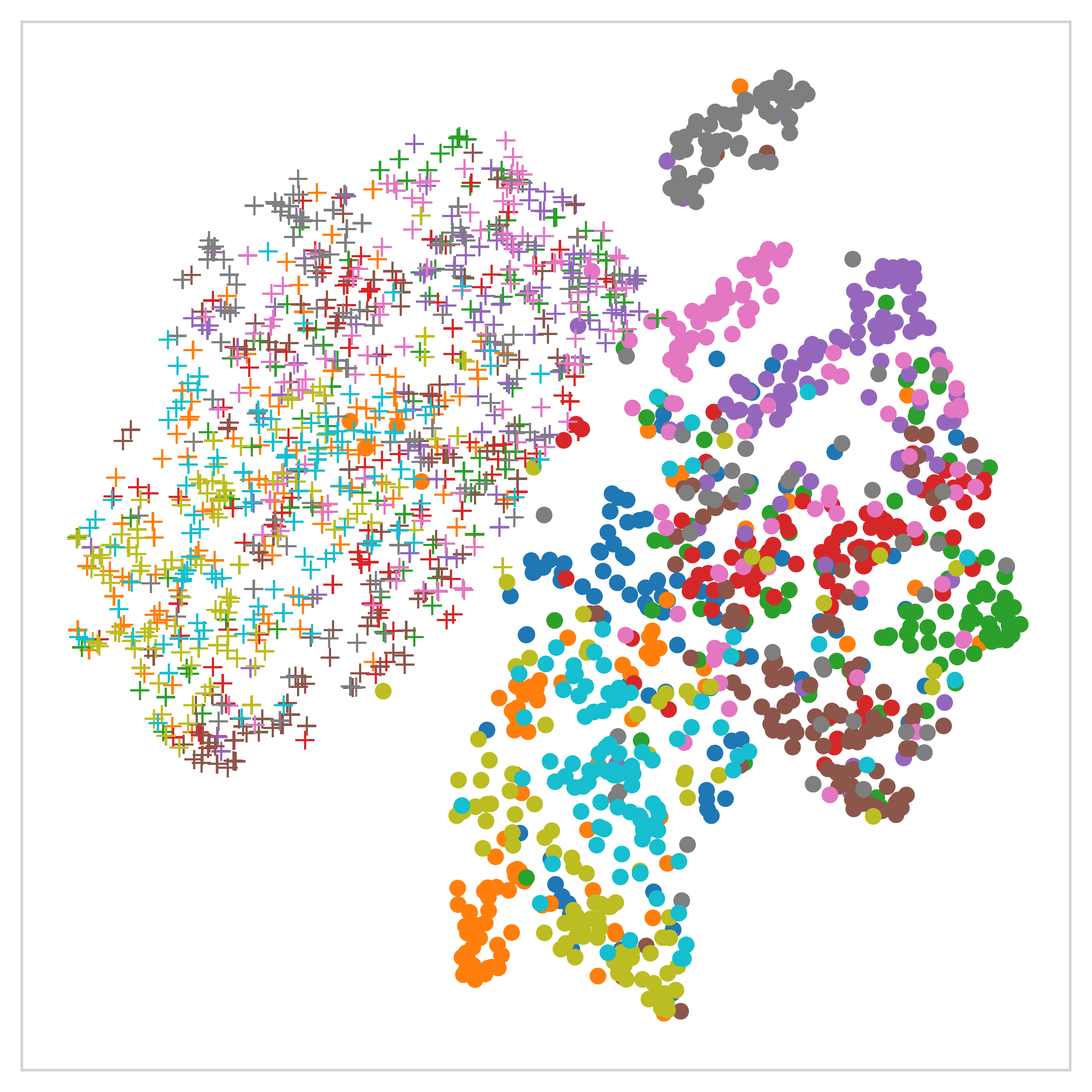}
	\caption{Stu (1st layer init, ep 0)}	
	\label{fig:cifar_tsne:g}
 	\end{subfigure}
 	\begin{subfigure}[]{0.195\textwidth}
	\includegraphics[width=\textwidth]{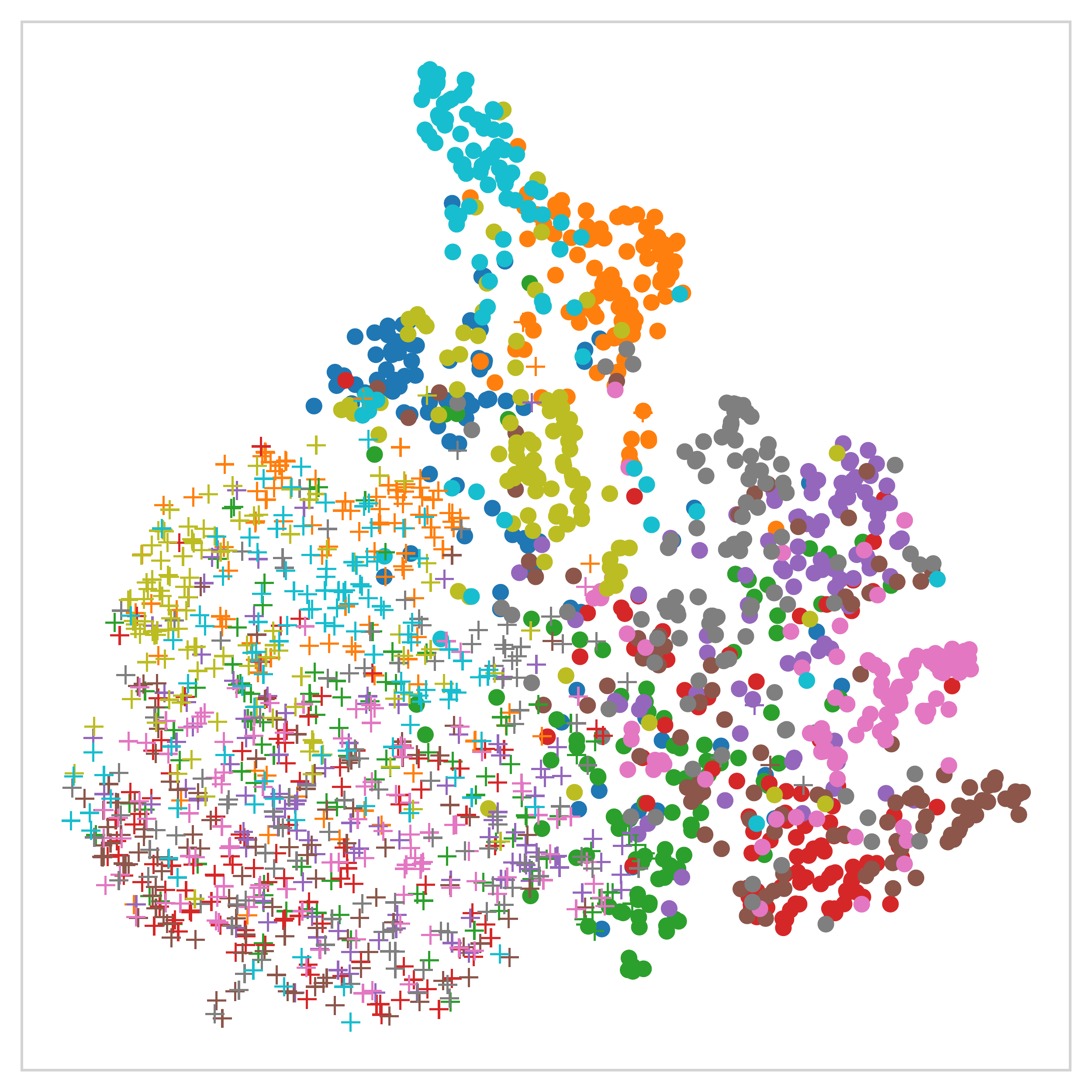}
	\caption{Stu (2nd layer init, ep 0)}	
	\label{fig:cifar_tsne:h}
 	\end{subfigure}
 	\begin{subfigure}[]{0.195\textwidth}
	\includegraphics[width=\textwidth]{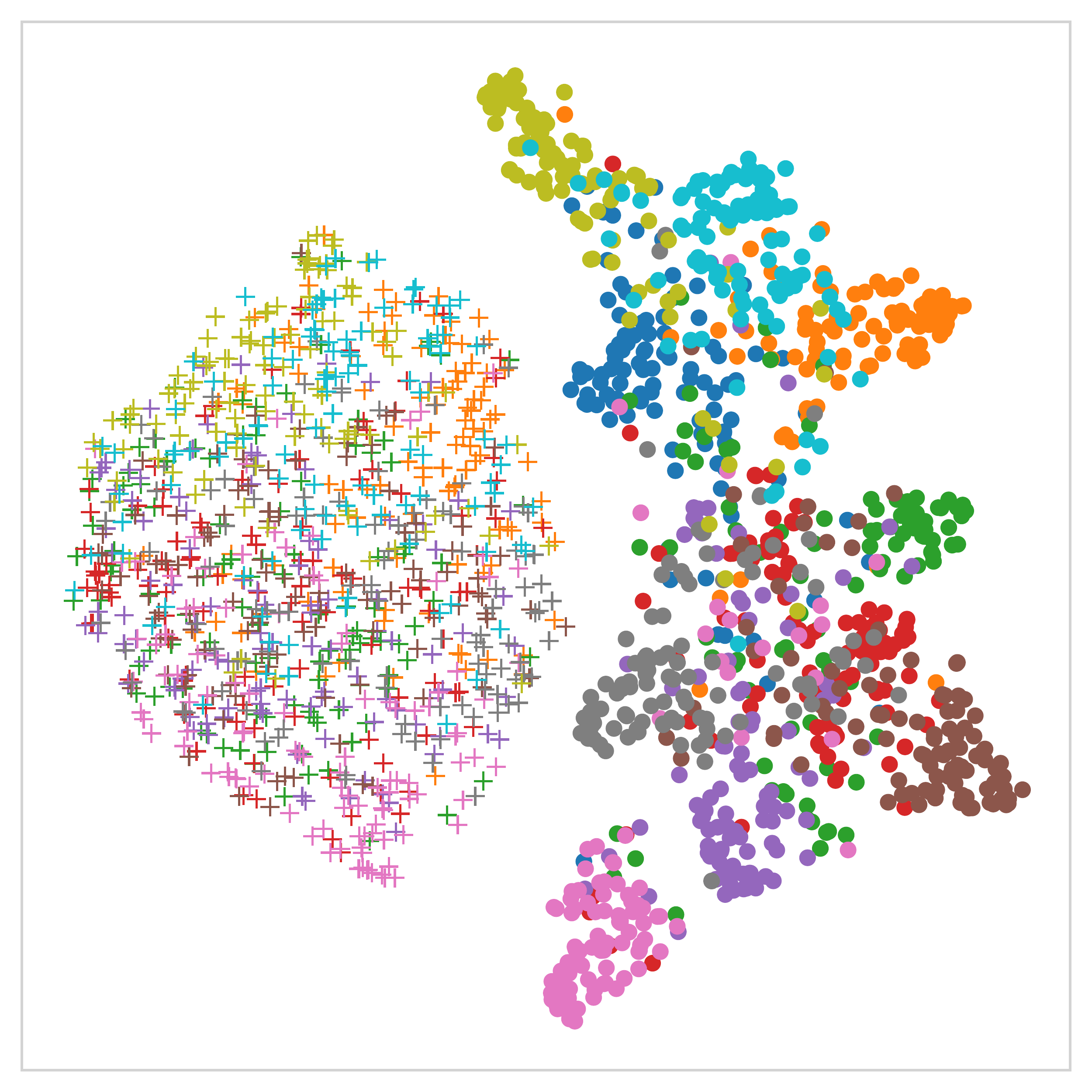}
    \caption{Stu (3rd layer init, ep 0)}	
	\label{fig:cifar_tsne:i}
 	\end{subfigure}
 	\begin{subfigure}[]{0.195\textwidth}
	\includegraphics[width=\textwidth]{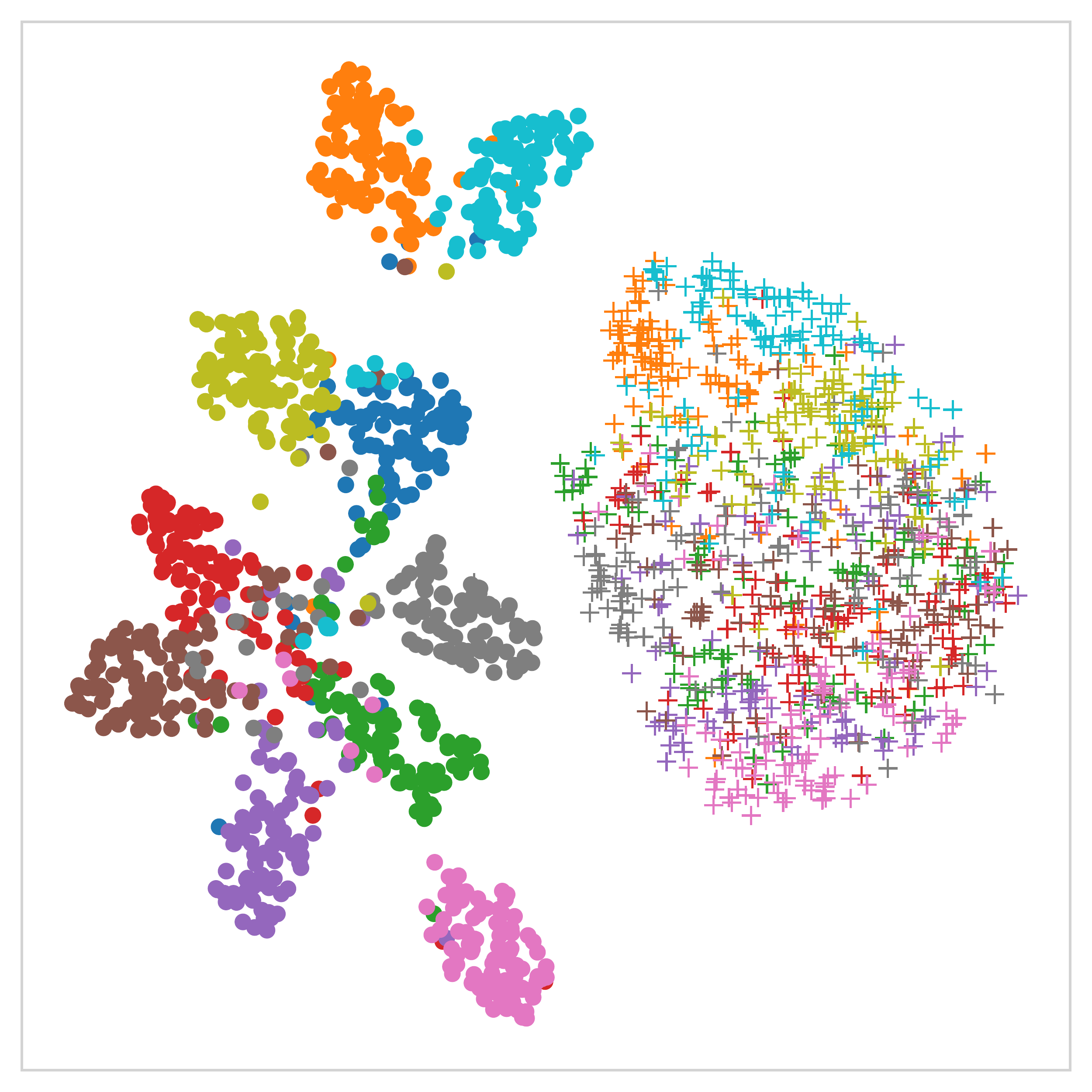}
    \caption{Stu (4th layer init, ep 0)}
	\label{fig:cifar_tsne:j}
\end{subfigure}
\caption{$t$-SNE visualization of penultimate features on CIFAR10 from \textit{Blended} attack. \textbf{Top}: the teacher model and student models at different training epochs  with adaptive layer-wise initialization. \textbf{Bottom}: student models at epoch 0 with different initialization strategies. Each color denotes a class. `$\circ$' are clean images and `$+$' the corresponding backdoor ones.}
    \vspace{-3mm}
    \label{fig:cifar_tsne}
\end{figure*}

\section{Experimental Details on Backdoor Defenses}
For all defenses, we train $100$ epochs. We adopt standard finetuning method using Stochastic Gradient Descent (SGD) with a momentum of $0.9$ and a weight decay of $0.0005$ with learning rate as $0.01$. We re-implement Fine-Pruning\footnote{\url{https://github.com/kangliucn/Fine-pruning-defense}} on ResNet18. As suggested in the Fine-Pruning~\cite{liu2018fine-pruning}, we prune the neurons in the last convolution layer. Since a residual block is integrated in ResNet, we actually prune neurons in the last two blocks. 
For NAD~\cite{li2021NAD}, we replicate the code\footnote{\url{https://github.com/bboylyg/NAD}} on ResNet18 and three attention maps are obtained from layer-$2$, layer-$3$ and layer-$4$ in PyTorch code. Other parameters are same as the released open-source code. For MCR~\cite{zhao2020MCR}, we replicate the code\footnote{\url{https://github.com/IBM/model-sanitization}} on ResNet18 and choose to find a path connection between a benign model and the original backdoor model. Similar to NAD, the benign model is obtained by applying standard finetuning on the original backdoor model after $10$ epochs.
We follow the open-source code\footnote{\url{https://github.com/csdongxian/ANP_backdoor}} of ANP~\cite{wu2021ANP}, and prune neurons by threshold. For I-BAU, we also use the released open-source code\footnote{\url{https://github.com/YiZeng623/I-BAU}}. For fair comparison, we train $100$ rounds instead of $5$ rounds in the original setting.

\section{Additional Qualitative Analysis}
In order to analyze our proposed method, we visualize penultimate features on CIFAR10 from Badnets attack in our paper. Here, we provide additional analysis on Blended attack. 

To analyze the effectiveness of knowledge distillation, we use $t$-SNE to visualize penultimate features across different training epochs and plot in the first row of Figure~\ref{fig:cifar_tsne}. The intra-class compactness and inter-class separability of clean samples reflect models' classification ability on clean samples. If backdoor samples are classified into clusters corresponding to their original labels, the model is clean and backdoor behavior has been removed. In Figure~\ref{fig:cifar_tsne:a}, we show the features of backdoor teacher model. The clean samples form 10 clusters, indicating that backdoor teacher model can predict labels of clean samples accurately. The backdoor samples form one single cluster distant from clean images. Consequently, backdoor teacher model behaves abnormally for backdoor images. The features after adaptive layer-wise initialization of student are shown in Figure~\ref{fig:cifar_tsne:b}. We can see that clean samples from same class still cluster together. Hence some benign knowledge are preserved in the student network. From Figure~\ref{fig:cifar_tsne:c} to Figure~\ref{fig:cifar_tsne:e}, we show features after training for 1, 2, 11 epochs respectively. We can observe that the clusters of clean samples become tighter and backdoor samples spread in the clusters of clean samples. The change of clusters reflects that benign knowledge is transferred into the student network gradually. Therefore, student model becomes a clean model without backdoor behavior.

In the second row of Figure~\ref{fig:cifar_tsne}, we visualize penultimate features of clean and backdoor samples from student model to analyze the effectiveness of adaptive layer-wise initialization. With uniform initialization and single-layer initialization, we can analyze qualitatively the influence of layer initialization from visualized characteristics of features. In order to keep the number of randomly initialized weights same, we get a uniform initialization ratio from our adaptive layer-wise initialization strategy. Figure~\ref{fig:cifar_tsne:f} shows the results of visualized features. Compare to Figure~\ref{fig:cifar_tsne:b},  Figure~\ref{fig:cifar_tsne:f} indicates that both clean samples and backdoor samples are scattered after uniform initialization of student model. Therefore, adaptive layer-wise initialization can preserve more benign knowledge than uniform initialization. From Figure~\ref{fig:cifar_tsne:g} to Figure~\ref{fig:cifar_tsne:j}, we show features of single-layer initialization of student model from first layer to fourth layer. When we initializing the lower layers e.g. first layer and second layer in Figure~\ref{fig:cifar_tsne:g} and Figure~\ref{fig:cifar_tsne:h}, the connection between trigger and target label is broken since backdoor samples and clean samples stay closer. However, benign knowledge is also ignored because clean samples do not form tight clusters corresponding to labels. When we initializing the higher layers e.g. third layer and fourth layer in Figure~\ref{fig:cifar_tsne:i} and Figure~\ref{fig:cifar_tsne:j}, more benign knowledge is preserved while the connection between trigger and target label is partially broken. Therefore, to obtain a trade-off between preserving benign knowledge and removing backdoor knowledge, the initialization ratios of lower layers should be smaller and the initialization ratios of higher layers should be larger. This provide evidences that our adaptive layer-wise initialization is reasonable.

\section{Results on ImageNet}
We also conduct experiments on a complex dataset: ImageNet. We choose $10$ classes from ImageNet to do attack and defense experiments. For each class, we split original training dataset into training dataset ($1000$ images) and validation dataset ($300$ images). Since ImageNet is a large dataset and training attack models requires more resources, we only choose Badnets, IAB, SIG and WaNet as attack models. Table~\ref{tab:imagenet} shows results. Our method is as good as the best existing methods, which depend on training labels. 

\renewcommand{\tabcolsep}{0.0cm}
\setlength{\aboverulesep}{1pt}
\setlength{\belowrulesep}{1pt}
\begin{table}[!h]
    \centering
    \footnotesize\scalebox{0.74}{
    \begin{tabular}{p{1.2cm}<{\centering}||p{0.6cm}<{\centering}p{0.6cm}<{\centering}||p{0.6cm}<{\centering}p{0.6cm}<{\centering}|p{0.6cm}<{\centering}p{0.6cm}<{\centering}|p{0.6cm}<{\centering}p{0.6cm}<{\centering}|p{0.6cm}<{\centering}p{0.6cm}<{\centering}|p{0.6cm}<{\centering}p{0.6cm}<{\centering}|p{0.6cm}<{\centering}p{0.6cm}<{\centering}||p{0.6cm}<{\centering}p{0.6cm}<{\centering}}
        \toprule

        \multirow{2}{*}{\tabincell{c}{Backdoor\\Attacks}} & \multicolumn{2}{c||}{Original} & \multicolumn{2}{c|}{FT} & \multicolumn{2}{c|}{FP} & \multicolumn{2}{c|}{MCR} & \multicolumn{2}{c|}{ANP} & \multicolumn{2}{c|}{NAD} & \multicolumn{2}{c||}{I-BAU} & \multicolumn{2}{c}{Ours}\\

        \cmidrule{2-17}
         & ASR & ACC & ASR & ACC & ASR & ACC & ASR & ACC & ASR & ACC & ASR & ACC & ASR & ACC & ASR & ACC \\
        \midrule
        Badnets & 100.0 & 77.4 & 0.4 & 78.0 & 99.6 & 77.4 & 4.7 & 76.6 & 99.8 & 71.4 & 1.6 & 78.8 & 1.8 & 71.2 & 0.2 & 78.0 \\ 
        IAB & 99.8 & 76.0 & 0.4 & 74.8 & 7.8 & 75.2 & 2.9 & 75.6 & 97.6 & 70.0 & 2.0 & 75.0 & 1.1 & 68.6 & 0.9 & 78.0 \\ 
        SIG & 91.8 & 79.2 & 0.0 & 77.6 & 0.4 & 75.8 & 29.1 & 74.6 & 91.8 & 79.2 & 0.0 & 78.8 & 0.0 & 71.2 & 0.2 & 79.6 \\ 
        WaNet & 98.7 & 79.8 & 5.8 & 78.0 & 21.3 & 78.8 & 1.1 & 75.8 & 98.7 & 79.8 & 2.7 & 78.4 & 2.2 & 73.4 & 8.2 & 79.8 \\ 
        \hline
        Mean & 97.6 & 78.1 & 1.7 & 77.1 & 32.3 & 76.8 & 9.4 & 75.7 & 96.9 & 75.1 & 1.6 & 77.8 & 1.3 & 71.1 & 2.4 & 78.9 \\  
        \bottomrule
    \end{tabular} }
    \vspace{-3mm}
    \caption{Defense results on backdoor models trained on ImageNet10.}
    \label{tab:imagenet}
\end{table}

\section{Experiments with different poison rates.}
We report results on different poison rates. We focus on BadNet attack with CIFAR10.
In the table~\ref{tab:poison_ratio_all2one}, our method performs well across different poison rate. We do observe an increase of ASR for large poison rate (20\%). The potential reason is that we focus on all-to-one attack setting; all triggered data are misclassified toward a single target class. This introduces bias toward the target class, especially for high poison rates. Such bias is inherited by the student model and can be hard to be mitigated by KD.

Table~\ref{tab:poison_ratio_all2all} reports all-to-all Badnets attack setting results. The results show that our method gets a better performance on high poison rates. 

\renewcommand{\tabcolsep}{0.0cm}
\setlength{\aboverulesep}{1pt}
\setlength{\belowrulesep}{1pt}
\begin{table}[!h]
    \centering
    \footnotesize\scalebox{0.74}{
    \begin{tabular}{p{1.2cm}<{\centering}||p{0.6cm}<{\centering}p{0.6cm}<{\centering}||p{0.6cm}<{\centering}p{0.6cm}<{\centering}|p{0.6cm}<{\centering}p{0.6cm}<{\centering}|p{0.6cm}<{\centering}p{0.6cm}<{\centering}|p{0.6cm}<{\centering}p{0.6cm}<{\centering}|p{0.6cm}<{\centering}p{0.6cm}<{\centering}|p{0.6cm}<{\centering}p{0.6cm}<{\centering}||p{0.6cm}<{\centering}p{0.6cm}<{\centering}}
        \toprule
        \multirow{2}{*}{\tabincell{c}{Poison\\Rates}}  & \multicolumn{2}{c||}{Original} & \multicolumn{2}{c|}{FT} & \multicolumn{2}{c|}{FP} & \multicolumn{2}{c|}{MCR} & \multicolumn{2}{c|}{ANP} & \multicolumn{2}{c|}{NAD} & \multicolumn{2}{c||}{I-BAU} & \multicolumn{2}{c}{Ours}   \\
        
        \cmidrule{2-17}
         & ASR & ACC & ASR & ACC & ASR & ACC & ASR & ACC & ASR & ACC & ASR & ACC & ASR & ACC & ASR & ACC \\
        \midrule
        1\% & 98.2 & 93.3 & 83.5 & 92.0 & 54.1 & 92.7 & 47.8 & 90.8 & 10.7 & 86.7 & 79.9 & 92.0 & 7.4 & 91.0 & 3.6 & 91.7 \\
        5\% & 99.6 & 92.5 & 48.4 & 91.5 & 50.3 & 92.1 & 24.5 & 90.3 & 1.6 & 85.2 & 47.4 & 91.4 & 1.7 & 91.0 & 5.0 & 91.3  \\ 
        10\% & 99.9 & 92.8 & 9.7 & 92.5 & 32.4 & 92.6 & 1.7 & 86.4 & 2.6 & 88.6 & 4.7 & 92.3 & 10.2 & 92.0 & 3.0 & 92.1  \\ 
        20\% & 100.0 & 88.6 & 6.8 & 89.5 & 89.6 & 90.2 & 4.4 & 88.2 & 3.1 & 85.4 & 1.6 & 89.3 & 2.0 & 89.0 & 19.3 & 88.6  \\ 
        \bottomrule
    \end{tabular} }
    \vspace{-3mm}
    \caption{Defense results on BadNet (all-to-one) with different poison rates.}
    \label{tab:poison_ratio_all2one}
\end{table}

\renewcommand{\tabcolsep}{0.0cm}
\setlength{\aboverulesep}{1pt}
\setlength{\belowrulesep}{1pt}
\begin{table}[!h]
    \centering
    \footnotesize\scalebox{0.74}{
    \begin{tabular}{p{1.2cm}<{\centering}||p{0.6cm}<{\centering}p{0.6cm}<{\centering}||p{0.6cm}<{\centering}p{0.6cm}<{\centering}|p{0.6cm}<{\centering}p{0.6cm}<{\centering}|p{0.6cm}<{\centering}p{0.6cm}<{\centering}|p{0.6cm}<{\centering}p{0.6cm}<{\centering}|p{0.6cm}<{\centering}p{0.6cm}<{\centering}|p{0.6cm}<{\centering}p{0.6cm}<{\centering}||p{0.6cm}<{\centering}p{0.6cm}<{\centering}}
        \toprule
        \multirow{2}{*}{\tabincell{c}{Poison\\Rates}}  & \multicolumn{2}{c||}{Original} & \multicolumn{2}{c|}{FT} & \multicolumn{2}{c|}{FP} & \multicolumn{2}{c|}{MCR} & \multicolumn{2}{c|}{ANP} & \multicolumn{2}{c|}{NAD} & \multicolumn{2}{c||}{I-BAU} & \multicolumn{2}{c}{Ours}   \\
        
        \cmidrule{2-17}
         & ASR & ACC & ASR & ACC & ASR & ACC & ASR & ACC & ASR & ACC & ASR & ACC & ASR & ACC & ASR & ACC \\
        \midrule
        1\% & 82.4 & 92.6 & 66.5 & 91.1 & 30.3 & 91.6 & 74.8 & 90.4 & 27.5 & 87.4 & 52.5 & 91.1 & 7.1 & 90.5 & 4.5 & 91.2\\
        5\% & 88.9 & 92.0 & 60.3 & 90.8 & 3.1 & 91.4 & 23.9 & 89.9 & 5.9 & 87.6 & 35.5 & 91.0 & 3.7 & 90.4 & 5.8 & 91.1\\
        10\% & 90.0 & 91.8 & 34.4 & 90.7 & 3.0 & 91.2 & 6.3 & 89.6 & 2.8 & 84.7 & 11.6 & 90.9 & 3.5 & 90.0 & 7.2 & 90.9\\
        20\% & 91.1 & 91.6 & 15.9 & 90.7 & 2.7 & 91.3 & 5.8 & 89.6 & 2.0 & 87.3 & 7.9 & 90.6 & 4.9 & 90.0 & 9.9 & 91.1\\
        \bottomrule
    \end{tabular} }
    \vspace{-3mm}
    \caption{Defense results on BadNet (all-to-all) with different poison rates.}
    \label{tab:poison_ratio_all2all}
\end{table}

{\small
\bibliographystyle{ieee_fullname}
\bibliography{egbib}
}

\appendix
\clearpage